\documentclass[journal]{IEEEtran}
\usepackage{booktabs}
\usepackage{amstext}
\usepackage{makecell}
\usepackage{multirow}
\usepackage{amssymb}
\usepackage{caption}

\usepackage[colorlinks=true,linkcolor=blue,citecolor=blue]{hyperref}

\ifCLASSINFOpdf
  \usepackage[pdftex]{graphicx}
\else
  \usepackage[dvips]{graphicx}
\fi
\usepackage{amsmath}
\usepackage{amsthm}
\newtheorem{definition}{DEFINITION}

\newtheorem{scenario}{SCENARIO}

\ifCLASSOPTIONcompsoc
 \usepackage[caption=false,font=normalsize,labelfont=sf,textfont=sf]{subfig}
\else
 \usepackage[caption=false,font=footnotesize]{subfig}
\fi
\begin{document}
%
\title{Continual Learning for Smart City: A Survey}
%
%
%

\author{Li Yang, 
        Zhipeng Luo*,
        Shiming Zhang,
        Fei Teng,
        and Tianrui Li~\IEEEmembership{Senior member,~IEEE,}%
\thanks{
Li Yang, Zhipeng Luo (the corresponding author), Shiming Zhang, Fei Teng, and Tianrui Li are with a) School of Computing and Artificial Intelligence, Southwest Jiaotong University, Chengdu 611756, China. b) Engineering Research Center of Sustainable Urban Intelligent Transportation, Ministry of Education. c) National Engineering Laboratory of Integrated Transportation Big Data Application Technology, Southwest Jiaotong University. d) Manufacturing Industry Chains Collaboration and Information Support Technology Key Laboratory of Sichuan Province, Southwest Jiaotong University, Chengdu 611756, Sichuan, P.R. China.
}%
\thanks{E-mails: yangli\_ef@my.swjtu.edu.cn, zpluo@swjtu.edu.cn, zhangshiming@my.swjtu.edu.cn, fteng@swjtu.edu.cn, trli@swjtu.edu.cn}%
\thanks{Manuscript received xxx; revised August xxx.}%
}

%
%

\markboth{Journal of \LaTeX\ Class Files,~Vol.~14, No.~8, August~2024}%
{Shell \MakeLowercase{\textit{et al.}}: Bare Demo of IEEEtran.cls for IEEE Journals}
%



\maketitle

\begin{abstract}
With the digitization of modern cities, large data volumes and powerful computational resources facilitate the rapid update of intelligent models deployed in smart cities.
Continual learning (CL) is a novel machine learning paradigm that constantly updates models to adapt to changing environments, where the learning tasks, data, and distributions can vary over time.
Our survey provides a comprehensive review of continual learning methods that are widely used in smart city development. 
The content consists of three parts: 
1) Methodology-wise. We categorize a large number of basic CL methods and advanced CL frameworks in combination with other learning paradigms including graph learning, spatial-temporal learning, multi-modal learning, and federated learning.
2) Application-wise. We present numerous CL applications covering transportation, environment, public health, safety, networks, and associated datasets related to urban computing.
3) Challenges. We discuss current problems and challenges and envision several promising research directions. We believe this survey can help relevant researchers quickly familiarize themselves with the current state of continual learning research used in smart city development and direct them to future research trends.

\end{abstract}


%
\IEEEpeerreviewmaketitle

\section{Introduction}
%
%
%
%

\IEEEPARstart{N}{owadays}, with advanced information technologies deployed citywide, large data volumes, powerful computational resources, and AI-based technologies are intelligentizing modern city development. Smart city research analyzes urban conditions and resident behaviors through urban data, aiming to enhance efficiency, sustainability, and quality of life for city residents.
Driven by fastly-iterated data collection and machine learning models, the development of a smart city also requires rapid and continuing progression.
To handle such a challenge, however, the standard practice is to purely retrain models with incrementally collected data, which can incur an affordable waste of time and computing resources.
Therefore, a more efficient solution called continual learning (CL) has recently drawn enormous attention from the machine learning community. In general, CL aims to help models rapidly learn new knowledge, which might come in the form of new data, new classes, or new domains.
The primary advantage of CL is that it only needs to \textit{update} models, \textit{as if} the new model has been retrained with both the previous and the new data. Such an update is usually low-key and can be done more frequently to achieve fast iterations.


\begin{figure}[tbp]
\centering
\includegraphics[width=3.4in]{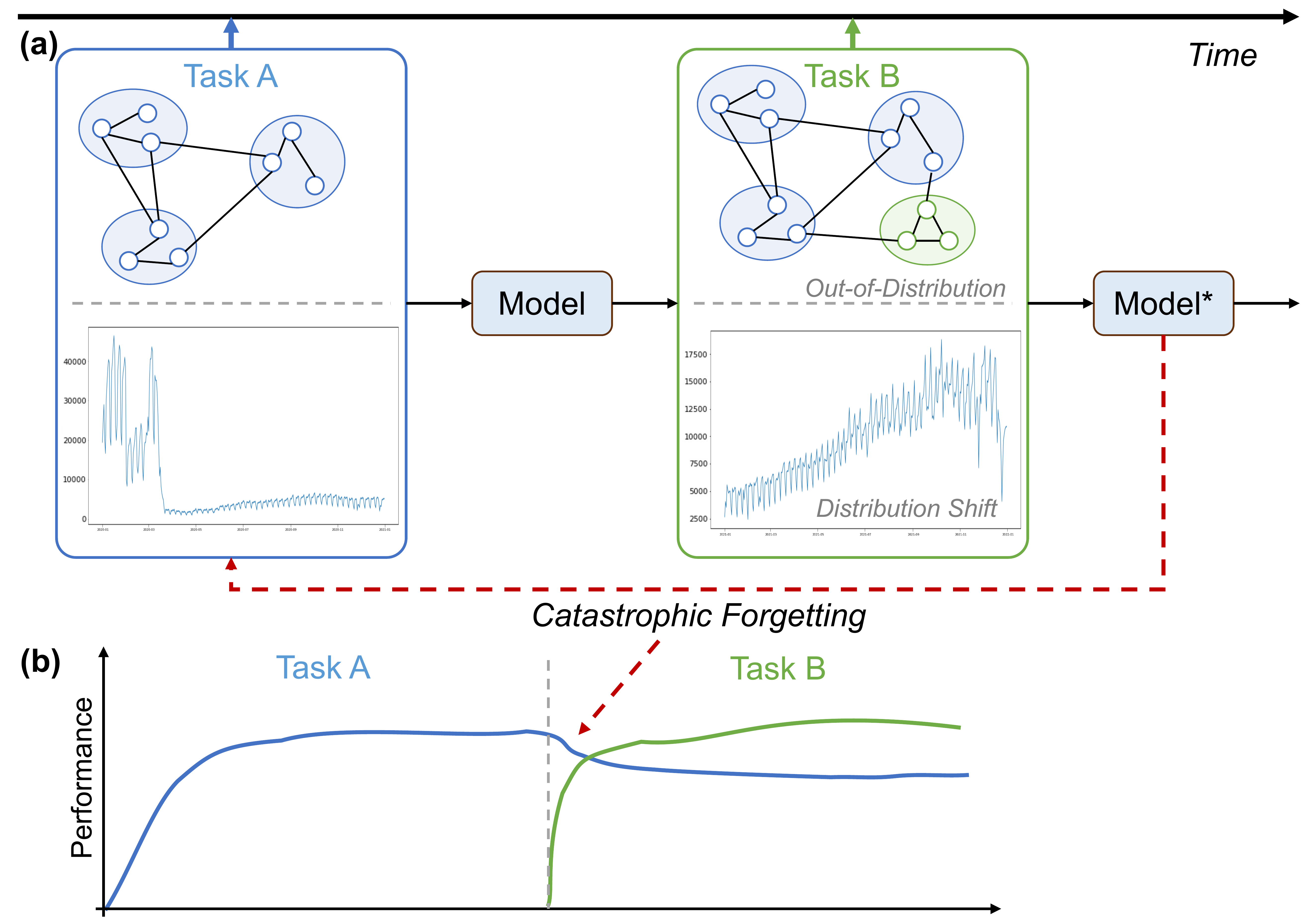}
\caption{An example of continual learning for graph learning. In \textbf{(a)}, a \textit{Model} is firstly trained based on the data from Task A and then updated by Task B as \textit{Model}*. Often, the new data distribution in Task B can be OOD. If the catastrophic forgetting problem is not handled well enough, the latest \textit{Model}* can have degenerated performance on the previous Task A, shown in \textbf{(b)}.}
\label{fig:introduction}
\end{figure}

The typical setting of continual learning is to learn a series of tasks sequentially.
In most CL scenarios, there is no universal assumption made about the data distribution among tasks. In fact, tasks can be generated from very different environments, so the new data can be out-of-distribution (OOD) or with the distribution shifted (DS) \cite{parisiContinualLifelongLearning2019,venThreeScenariosContinual2019}.
The difference in data distributions poses a critical challenge in CL, that is, how the model can continually learn new knowledge while not forgetting what it has learned. It is conversely rephrased as the Catastrophic Forgetting (CF) problem \cite{mccloskeyCatastrophicInterferenceConnectionist1989} if models continually forget during learning. The key to dealing with this problem is to balance the \textit{plastic learning} ability and the \textit{stable memory} ability. The former refers to a model's ability of continually acquiring new knowledge, while the latter refers to memorizing learned knowledge.
Fig.\ref{fig:introduction} illustrates a graph learning example where the graph, say, representing a city road network, can expand with time. A good model should be able to remember enough features or patterns of the previous graph structure (Task A) and then keep learning the new ones (Task B).

\begin{figure}[tbp]
\centering
\includegraphics[width=3.4in]{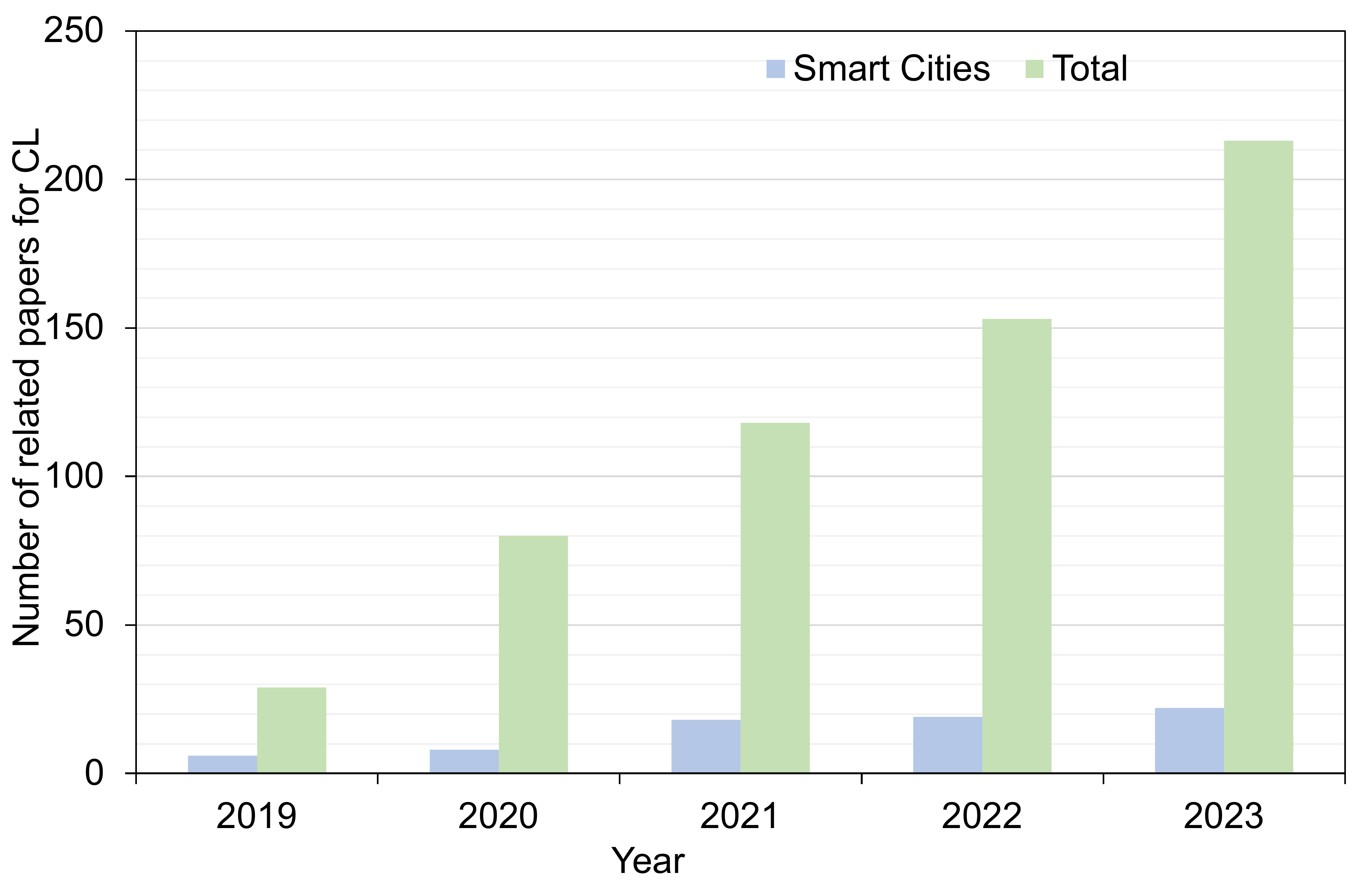}
\caption{The trend of continual learning research and its application to smart city research. Statistics are from Google Scholar over the past five years.}
\label{fig:paperNumber}
\end{figure}

Recently, we have experienced a surge of continual learning research in the machine learning community. 
We performed an exhaustive search and count of CL-related papers published within the past five years at major conferences including CVPR, ACL, ICML, ICLR, NeurIPS, and journals like IEEE TPAMI, TKDE, and IJCV. The figure shows that the research activities on CL increased significantly from 29 publications in 2019 to more than 200 in 2023, resulting in a total increase of more than 10 times.
Early works began with iCaRL\cite{rebuffiICaRLIncrementalClassifier2017}, EWC\cite{kirkpatrickOvercomingCatastrophicForgetting2017}, and LwF\cite{liLearningForgetting2018} in 2017. Sooner, the computing vision community became the major field where CL prospered. 
A lot of architecture-based, replay-based, and regularization-based methods were proposed and achieved remarkable results in different CL settings like incremental data, classes, domains, online learning, etc.
Such rapid growth of CL research provides an initial basis for studying more complex and practical problems in urban computing \cite{zhengUrbanComputingConcepts2014}. 
During the same period, CL was gradually studied in various applications in smart cities, including transportation, environment, public health, public safety, public networks, auto-vehicles, and robots. We counted publications related to both CL and smart cities and plotted the statistics in Fig.\ref{fig:paperNumber}. This implies that smart city related CL works are in a strong uptrend and many promising challenges remain open.

\textbf{Related Surveys.} The number of existing surveys on different aspects of continual learning isn't small. For example, the most recent work by
Wang et al. \cite{wangComprehensiveSurveyContinual2023} completed a very comprehensive review of the latest CL work as of 2023. Other reviews mainly focused on computer vision and natural language processing \cite{parisiContinualLifelongLearning2019, belouadahComprehensiveStudyClass2021, delangeContinualLearningSurvey2022, keContinualLearningNatural2023, zhouDeepClassIncrementalLearning2023}. Specifically, Belouadah et al. \cite{belouadahComprehensiveStudyClass2021} defined six expected properties of incremental learning algorithms and proposed a universal evaluation framework; De Lange et al. \cite{delangeContinualLearningSurvey2022} proposed a widely recognized classification method for continual learning; Ke and Liu \cite{keContinualLearningNatural2023} provided a classification of CL methods in Natural Language Processing; Parisi et al. \cite{parisiContinualLifelongLearning2019} presented the main challenges of continual learning from the perspective of biological principles; Zhou et al. \cite{zhouDeepClassIncrementalLearning2023} provided detailed experiments on the differences and advantages of various CL methods in the vision field. There have also been many reviews on graph neural networks that have received widespread attention \cite{yuanContinualGraphLearning2023, febrinantoGraphLifelongLearning2023}. In other fields, Shaheen et al. \cite{shaheenContinualLearningRealWorld2022} provided a review of applications in real-world automation systems; Lesort et al. \cite{lesortContinualLearningRobotics2019} investigated the CL methods and applications to robotics; and finally, Zhang and Kim \cite{zhangSurveyIncrementalUpdate2023} analyzed the CL methods and applications to recommendation systems.
Despite the rich number of surveys, none has particularly focused on the continual learning work done for urban computing or smart cities, where many important studies should not be overlooked.
Therefore, our survey is devoted to filling this gap by reviewing the recent edge-cutting CL work done for smart cities.

\textbf{Contributions.} The contributions of our survey are summarized as follows:
\begin{itemize}
    \item To the best of our knowledge, this is the first survey that reviews the latest continual learning work for smart cities. We summarize the advances from both application and methodology perspectives based on very rich literature.
    \item We categorize the major applications and their CL problem formulations in smart city scenarios. Meanwhile, we also list out numerous public datasets associated.
    \item We introduce various advanced continual learning frameworks common to smart cities that integrate other learning paradigms such as graph learning, temporal Learning, spatial-temporal learning, multi-modality learning, and federated learning. 
    \item Lastly, we discuss existing challenges of continual learning research for smart cities and envision several promising future directions
\end{itemize}

\textbf{Organization.} The rest of this survey is organized as follows. In Section \ref{sec-background}, we introduce the background of continual learning, including the basic task setting, common scenarios of CL in smart cities, and methods for overcoming catastrophic forgetting. In Section \ref{sec-application}, we summarize various applications of CL to smart cities and elaborate on how CL is used in different scenarios. In Section \ref{sec-framework}, we present advanced CL frameworks in combination with other learning paradigms. In Section \ref{sec-discussion}, we analyze the current open problems and challenges and suggest several few directions.
Finally, we conclude our survey in Section \ref{sec-conclusion}.


\textbf{\begin{figure*}[htbp]
\centering
\includegraphics[width=7.2in]{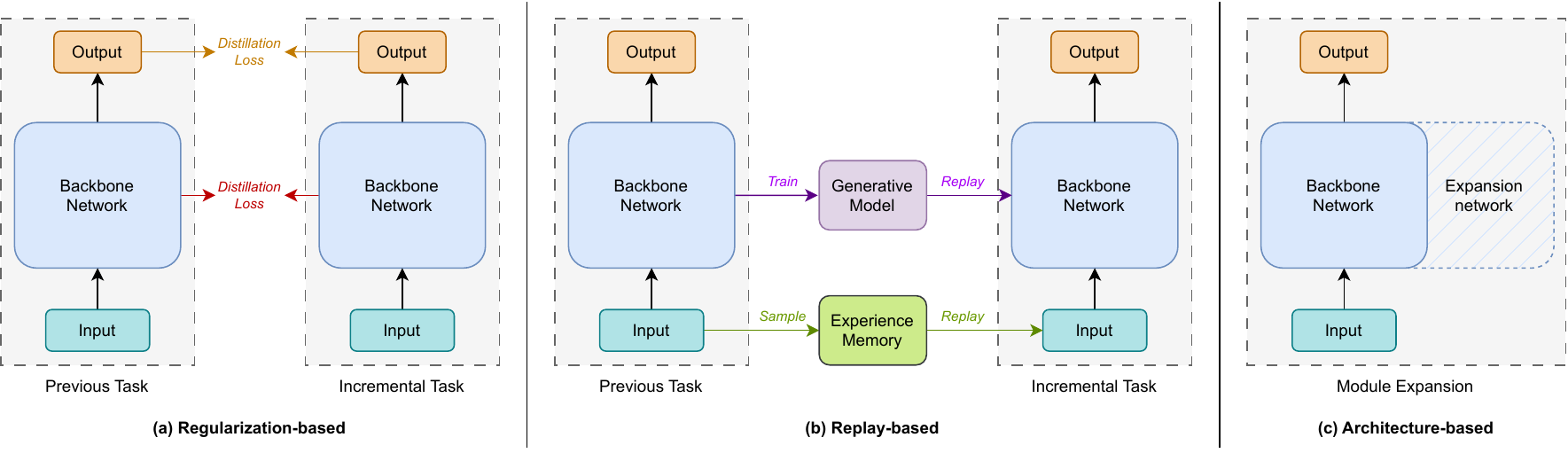}
\caption{A taxonomy of common continual learning methods to overcome catastrophic forgetting. \textbf{(a). Regularization-based methods} use additional regularization terms to prevent the important model parameters from deviating too much during sequential training. \textbf{(b). Replay-based methods} maintain a memory buffer or a generative model to replay the past data when training with new data. \textbf{(c). Architecture-based methods} memorize past knowledge by using previous network architecture and learn new knowledge by expanding the network's architecture. The expanded network can be neuron-level or module-level.}
\label{fig:methods}
\end{figure*}
}

\section{Background}
\label{sec-background}
We start by providing a brief introduction to smart cities and continual learning. We will cover the basic settings, scenarios, and a method taxonomy of continual learning. Also, we present two ways of categorization of CL methods applied to smart cities - Table \ref{tab:scenarios} is based on different CL scenarios while Table \ref{tab:application} is based on different smart application fields of smart cities.

\subsection{Smart Cities}
A smart city is a technologically modern urban area that relies on information and communication technology (ICT) as its technical backbone and uses sensors and user devices deployed throughout the city to capture real-time data on the urban condition. This infrastructure enables enhanced urban management and crisis mitigation through coordinated planning and behavioral guidance. The overarching objective of smart cities is to make cities and human settlements inclusive, safe, resilient, and sustainable\footnote{https://www.un.org/sustainabledevelopment/cities/}. With the ongoing advancement of the Internet of Things, big data, and artificial intelligence, smart cities are progressing toward greater digitization, intelligence, and sustainable development. Hence, we are witnessing the rapid development of smart city applications across various fields, including transportation, environmental protection, public safety, healthcare, network management, urban robotics, and many others.

\subsection{Task Settings and Scenarios of Continual Learning}
Continual learning, sometimes also known as lifelong learning, has been defined differently in the literature, and there is no consensus on a very accurate definition. Nevertheless, we can define some of the essential elements necessary for our discussion. See below.

\begin{definition}
    \textbf{Base Model}. The ultimate goal is to learn a supervised machine learning model $f: \mathcal{X} \rightarrow \mathcal{Y}$ that maps an input domain $\mathcal{X}$ to an output domain $\mathcal{Y}$. Often, $f$ is a parametric model, and we parameterize it by $\theta$ as $f_\theta$. In most CL studies, $f_\theta$ is a neural network as it has both strong plastic and stable abilities. $f_\theta$ is continually trained through a sequence of tasks.
\end{definition}

\begin{definition}
    \textbf{Sequential Tasks}. Suppose there is a series of tasks coming sequentially for a model to learn with. Each task has a task identifier $\tau \in \{1, \dots, \mathcal{T} \}$ and a corresponding (labeled) dataset $\mathcal{D}_\tau = \{ X_\tau, Y_\tau \}$. The dataset further consists of a training set $\mathcal{D}_\tau^{train}$ and a testing set $\mathcal{D}_\tau^{test}$. $\mathcal{T}$ can be either finite or infinite. The data $\mathcal{D}_\tau$ are assumed to be sampled from an unknown distribution $P_\tau(\mathcal{X},\mathcal{Y})$, and we can informally say that $P_\tau(\mathcal{X},\mathcal{Y})$ represents the knowledge of task $\tau$.
\end{definition}


\begin{definition}
    \textbf{Continual Learning}. Continual Learning is a process that trains a base model $f_\theta$ through sequential tasks. The typical setting is that when task $\tau$ comes, the model has been trained with the previous $\tau-1$ tasks, denoted by $f_\theta^{(\tau-1)}$; moreover, the model only has access to the data of the current task $\tau$, with which it is updated as $f_\theta^{(\tau)}$. The goal of CL is to have the final model $f_\theta^{(\mathcal{T})}$ generalize well on all the tasks' underlying distributions. To practically measure the generalizability, we test the model on all tasks' test data, and a good model should achieve a low total loss, as shown in Equation (\ref{eqn-avgloss}).
    \begin{equation}
        \label{eqn-avgloss}
        \theta^* =\underset{\theta}{\mathrm{argmin}} \frac{1}{\mathcal{T}} \sum_{\tau=1}^{\mathcal{T}} \mathcal{L} (f_\theta^{(\mathcal{T})}(X_\tau^{test}), Y_\tau^{test} )
    \end{equation}
\end{definition}

\begin{table*}[htbp]
\centering
\caption{Continual learning scenarios and related work in smart cities.}
\label{tab:scenarios}
\begin{tabular}{@{}lllc@{}}

\toprule
\textbf{Scenario} & \textbf{Data Distribution} & \textbf{Task Identifiability} & \textbf{Publications} \\ 
\midrule 
TIL\cite{venThreeScenariosContinual2019} 
    & \makecell{
    $\mathcal{X}_i = \mathcal{X}_j, \mathcal{Y}_i = \mathcal{Y}_j$ \\
    $P_i(\mathcal{X}, \mathcal{Y}) \ne P_j(\mathcal{X}, \mathcal{Y}) $
    } 
    & $\tau$ is available
    & \makecell{
    R2C-SVR-IL\cite{xiaoShorttermTrafficVolume2019},
    TF-Net\cite{yuIncrementalLearningBased2021},
    IL-TFNet\cite{shaoTrafficFlowPrediction2021},
    FpC\cite{lanzaUrbanTrafficForecasting2023},
    TrafficStream\cite{chenTrafficStreamStreamingTraffic2021},
    STKEC\cite{wangKnowledgeExpansionConsolidation2023},
    \\
    PAVEMENT\cite{maipraditPAVEMENTPassingVehicle2022},
    IKASL\cite{bandaragodaTrajectoryClusteringRoad2019},
    ISVM\cite{sunInternetTrafficClassification2018}
    CEDUP\cite{wuCEDUPUsingIncremental2023},
    SARDINE\cite{dasSARDINESelfAdaptiveRecurrent2020},
    LIL\cite{luLILLightweightIncremental2022},
    \\
    MCLNP\cite{wangUncertaintyEstimationNeural2022}, 
    SEIRD-IL\cite{camargoIncrementalLearningApproach2022},
    FIRF\cite{huNovelFeatureIncremental2019},
    HAR-IL\cite{younanDeepIncrementalLearning2022},
    Solver-CM\cite{guptaContinualLearningMultivariate2021},
    TTD\cite{yinContinualLearningAttentive2023},
    \\
    ISVM\cite{sunInternetTrafficClassification2018},
    ILAs\cite{eldhaiTrafficClassificationBased2022},
    KPGNN\cite{caoKnowledgePreservingIncrementalSocial2021},
    LL-ED\cite{yuLifelongEventDetection2021},
    FinEvent\cite{pengReinforcedIncrementalCrossLingual2023},
    GraphSAIL\cite{xuGraphSAILGraphStructure2020},
    \\
    GDumb\cite{ahrabianStructureAwareExperience2021},
    IGC\cite{dingCausalIncrementalGraph2022},
    STS-Rec\cite{amiratIncrementalTreebasedSuccessive2023},
    SACDLD\cite{wangGraphStructureAware2021},
    FIRE\cite{xiaFIREFastIncremental2022},
    FMHP\cite{cuiEventbasedIncrementalRecommendation2023},
    \\
    GCMTP\cite{maContinualMultiAgentInteraction2021},
    D-GSM\cite{linContinualInteractiveBehavior2023},
    SCL-PP\cite{knoedlerImprovingPedestrianPrediction2022},
    CLTP-MAN\cite{yangContinualLearningbasedTrajectory2022},
    CL-SGR\cite{wuContinualPedestrianTrajectory2023},
    AirLoop\cite{gaoAirLoopLifelongLoop2022},
    \\
    DVS-IL\cite{lunguFastEventdrivenIncremental2019},
    ISTMM\cite{kanazawaIncrementalLearningSpatialTemporal2019},
    CAA-HRI\cite{maharjanContinualLearningAdaptive2022},
    TCBLS\cite{caoIncrementalLearningRemaining2023}
    }
\\ \midrule 
DIL\cite{venThreeScenariosContinual2019} 
    & \makecell{$\mathcal{X}_i \ne \mathcal{X}_j, \mathcal{Y}_i = \mathcal{Y}_j$ \\
    $P_i(\mathcal{X}_i, \mathcal{Y}_i) \ne P_j(\mathcal{X}_j, \mathcal{Y}_j) $ } 
    & $\tau$ is optional
    & \makecell{
    FLCB\cite{gaoForgetLessCount2023},
    FedSTIL\cite{zhangSpatialTemporalFederatedLearning2023},
    DILRS\cite{ruiDILRSDomainIncrementalLearning2023},
    DENet\cite{wangDomainincrementalLearningFire2023},
    DR-EMR\cite{vijayaraghavanLifelongKnowledgeEnrichedSocial2021}
    }\\ \midrule 
CIL\cite{venThreeScenariosContinual2019} 
    & \makecell{ $\mathcal{Y}_i \cap \mathcal{Y}_j = \emptyset$  \\
    $P_i(\mathcal{X}_i, \mathcal{Y}_i) \ne P_j(\mathcal{X}_j, \mathcal{Y}_j) $
    }
    & $\tau$ is unavailable
    & \makecell{
    IEL\cite{hanPrototypeEnhancementBasedIncremental2023},
    CITS\cite{xueliClassIncrementalTemporalSpatial2020},
    Solver-CM\cite{guptaContinualLearningMultivariate2021},
    Dynamic-IL\cite{veerakumarDynamicIncrementalLearning2023},
    \\
    iCarl+\cite{bovenziFirstLookClass2021},
    KCN\cite{caoIncrementalEventDetection2020},
    EMP\cite{liuIncrementalPromptingEpisodic2022}
    }
\\  \midrule 
OCL\cite{aljundiGradientBasedSample2019} 
    & \makecell{Based on TIL, DIL, or CIL} 
    & $\tau$ is optional
    & \makecell{
    STMP\cite{nallaperumaOnlineIncrementalMachine2019},
    ArcVideo\cite{zhangDataEfficientContinuousLearning2022},
    GWR\cite{tenzerLearningCitywidePatterns2022},
    WISDOM\cite{xuSpatioTemporalMultiTaskLearning2021},
    ANNAIL\cite{farooqNovelAdaptiveDeep2020},
    \\
    DIM-BLS\cite{xiaoBaselineModelTraining2021},
    CSTWPP\cite{huVeryShorttermSpatial2020},
    HyperHawkes\cite{dubeyContinualLearningTimetoEvent2022},
    oHIML\cite{zhaoSpatioTemporalEventForecasting2021},
    DEGC\cite{heDynamicallyExpandableGraph2023},
    iEA\cite{hanIETARobustScalable2023}
    }
\\
\bottomrule
\end{tabular}
\end{table*}

The above definition of continual learning is quite general, and various specific scenarios can extend from it \cite{venThreeScenariosContinual2019, wangComprehensiveSurveyContinual2023}. Below we introduce some common scenarios, which are distinguished according to their task identifiability and data distribution. Specifically, task identifiability means whether the task id $\tau$ is available during the training or testing phase; and data distributions $P_\tau(\mathcal{X},\mathcal{Y})$, or even the domains $\mathcal{X}$ and $\mathcal{Y}$, may differ from task to task. We summarize these scenarios and their publications related to smart cities in Table \ref{tab:scenarios}.

\begin{scenario}
    \textbf{Task-incremental learning (TIL)}. In the TIL scenario, the base model can distinguish tasks by accessing their task identifiers in both the training and testing phases. The main challenge of TIL is to handle potentially different data distributions $P_\tau(\mathcal{X},\mathcal{Y})$ among tasks. And the domains  $\mathcal{X}$ and $\mathcal{Y}$ usually remain unchanged.
\end{scenario}

TIL is the most fundamental scenario and is widely studied in the area of smart cities, shown in Table \ref{tab:scenarios}. Typical examples are TrafficStreem~\cite{chenTrafficStreamStreamingTraffic2021} and STKEC~\cite{wangKnowledgeExpansionConsolidation2023} that study traffic flow prediction problems in continually expanded cities. The year naturally becomes the task ID and the traffic flow in each new year has a large distribution shift. Accordingly, a base prediction model is updated yearly to adapt to the changes in traffic flow.

\begin{scenario}
    \textbf{Domain-incremental learning (DIL)}. In the DIL scenario, the key problem is to deal with potentially different input domains $\mathcal{X}$, while the output domain $\mathcal{Y}$ usually remains unchanged. The task identifiers are optional.
\end{scenario}

DIL is often used to address environmental changes in smart cities. A typical application is remote sensing \cite{ruiDILRSDomainIncrementalLearning2023, wangDomainincrementalLearningFire2023}, where a model aims to identify city construction or natural disasters. However, with the changing of air condition, sensor lifetime, complex background, etc., the domain shifts between tasks can lead to severe catastrophic forgetting. In DIL, the model needs to learn on various domains with sequential tasks.

\begin{scenario}
    \textbf{Class-incremental learning (CIL)}. In the CIL scenario, the CL model should predict new classes in forthcoming tasks, and in some extreme settings, the number or the name of new classes is even not informed in the testing phases. Hence, the output domains $\mathcal{Y}$ can change. Meanwhile, the base model is unable to distinguish which task the new classes belong to through task identifiers.
\end{scenario}

The CIL scenario is a more recent research hotspot in the vision community, but it has not been fully investigated in smart city research. To name one of the few works, IEL \cite{hanPrototypeEnhancementBasedIncremental2023} proposed an incremental urban garbage classification setting. With new tasks, the base model continues to identify new garbage classes to achieve good performance on every task.

\begin{scenario}
    \textbf{Online Continual Learning (OCL)}. In the OCL scenario, the data of each task come in a smaller granularity, in streaming instances or batches. OCL is built on the above three scenarios in an online fashion. 
\end{scenario}

Online learning is a common setting in smart cities, such as traffic management system~\cite{nallaperumaOnlineIncrementalMachine2019} and recommendation system~\cite{heDynamicallyExpandableGraph2023}. Note that in some research, the term streaming data is used to denote online data, while in other cases, the term means incremental data.

\subsection{Continual Learning Methods}

There are many categories of continual learning methods, and here we introduce three common types. They are regularization-based, replay-based, and architecture-based methods, shown in Fig. \ref{fig:methods}.

\subsubsection{Regularization-based methods} 
Regularization-based methods are characterized by adding regularization terms that explicitly control the model's plastic and stable abilities. Usually, models trained by such methods have relatively stable memory but have limited capacity for learning new knowledge.
A typical implementation is to add a secondary penalty to the loss function, punishing the model's important parameters for large deviations during its continual learning process. 
To name a few, EWC~\cite{kirkpatrickOvercomingCatastrophicForgetting2017} was the first proposed work to reduce catastrophic forgetting (CF) by constraining important parameters. The importance was measured by the Fisher information matrix. SI~\cite{zenkeContinualLearningSynaptic2017} calculated the importance of parameters online during the model training phase. The sensitivity of the cumulative loss function to the change of each parameter during the training process is used as the estimated importance. MAS~\cite{aljundiMemoryAwareSynapses2018} used unlabeled samples to estimate the sensitivity of a neural network to parameter changes as an estimate of parameter importance.

\subsubsection{Replay-based methods}

Another more effective way to overcome catastrophic forgetting is to memorize a small portion of previous data when training with new data. Replay-based methods typically maintain a memory buffer or a generative model to \textquotedblleft replay\textquotedblright\ the past experience.
For example, Rebuffi et al.~\cite{rebuffiICaRLIncrementalClassifier2017} proposed for the first time the incremental classification and representation learning method iCaRL based on data replay. After learning each task, this method saves a small number of samples for each category for subsequent training. As the model can still see the saved representative data, the CF issue can be partly alleviated. Because of such an advantage, data replay techniques have been frequently used in numerous CL methods.

\subsubsection{Architecture-based methods}

Neural networks have very flexible architectures, so people have proposed a new CL approach that can achieve zero forgetting, which is to dynamically expand the network. Usually, the network's parameters are divided into parts that fit different tasks, so different knowledge is memorized without interfering. When necessary, the network can expand to allow learning new tasks.
Yan et al.~\cite{yanDynamicallyExpandableRepresentation2021} proposed a dynamically expandable representation learning (DER) method with a modular deep classifier network containing a super feature extractor network and a linear classifier. Specifically, the super feature extractor network comprises multiple feature extractors of different sizes, adjusted for each incremental step. When faced with new classes, DER extends the network with new feature extractors while freezing the previous ones. Finally, features from all extractors are concatenated for class prediction.


\subsection{Evaluation Metrics}
The performance of continual learning methods should be evaluated from global-local and forgetting-memory perspectives. We introduce below some common evaluation metrics frequently used in the CL literature.

\textbf{Average Accuracy} (AA) \cite{chaudhryRiemannianWalkIncremental2018a} measures the classification accuracy of the base model on both the past and the present tasks. The average accuracy of the current task $\tau$ is defined as follows:
\begin{equation}
    \label{eqn-AA}
    A_\tau = \frac{1}{\tau}\sum_{k=1}^{\tau} a_{\tau, k}
\end{equation}
where $ a_{\tau, k} $ denotes the classification accuracy on the testing set $ \mathcal{D}^{test}_k $ of task $k$ after training on task $\tau$. AA suggests the global performance of the model on all the trained datasets $\{ \mathcal{D}_1, \mathcal{D}_2, \cdots, \mathcal{D}_\tau \}$, but little does it reflects the local tasks and forgetting.

\textbf{Forgetting Measure} (FM) \cite{chaudhryRiemannianWalkIncremental2018a} measures a model's forgetting on all previous tasks after training on the current task. The average forgetting of the current task $\tau$ is defined as follows:
\begin{equation}
    \label{eqn-FM}
    F_\tau = \frac{1}{\tau-1}\sum_{k=1}^{\tau-1} f_k^\tau
\end{equation}
where $f_k^\tau$ denotes the difference between the maximum accuracy learned from all previous tasks and the accuracy after being updated by the current task. The forgetting at task $k$ after the CL model has been learned from the current task $\tau$ is defined as:
\begin{equation}
    \label{eqn-f}
    f_k^\tau = \max_{i \in \{1, \cdots,\tau-1 \}} a_{i,k} - a_{\tau,k}, \quad \forall\ k < \tau
\end{equation}
Apparently, large FM implies more forgetting.

\textbf{Backward Transfer} (BWT) \cite{lopez-pazGradientEpisodicMemory2017} measures the influence on how much the current learning task $\tau$ affects the historical task $k$ ($k < \tau$). BWT calculates the mean of the accuracy differences of task $k$ between before and after the CL model trained on the $\mathcal{D}_{\tau}^{train}$:
\begin{equation}
    \label{eqn-BWT}
    \mathrm{BWT_\tau } = \frac{1}{\tau - 1}\sum_{k=1}^{\tau - 1} a_{\tau,k} - a_{k,k} 
\end{equation}
A positive BWT indicates that the performance on task $t$ increases after the CL model has trained on task $\tau$. On the contrary, a large negative BWT means that forgetting happens on task $k$ after learning task $\tau$. 

\textbf{Forward Transfer} (FWT) \cite{lopez-pazGradientEpisodicMemory2017} measures the influence of the current learning task $\tau$ affecting the future task $k$ ($k > \tau$). FWT calculates the mean of the accuracy differences of task $k$ between the zero-shot learning and the randomly initialized network on the testing set $\mathcal{D}_{k}^{test}$. Here, the accuracy performance from the randomly initialized network is denoted as $b_k$:
\begin{equation}
    \label{eqn-FWT}
    \mathrm{FWT_\tau } = \frac{1}{\tau - 1}\sum_{k=2}^{\tau} a_{k-1,k} - b_k 
\end{equation}
A positive FWT shows that the CL model can transfer the knowledge from the preceding task to the current task, that is, the task $k-1$ can improve the performance on task $k$.

In addition to the above metrics, there are many other ones to evaluate CL methods from different perspectives \cite{wangComprehensiveSurveyContinual2023, keContinualLearningNatural2023}. We will provide a detailed explanation of the metrics closely related to smart cities in the following sections.


\section{Applications}
\label{sec-application}

In this section, we discuss task-specific challenges regarding the application of continual learning to smart cities. The literature we surveyed covers a wide range of areas. As shown in Tab. \ref{tab:application}, we include transportation, environment, public health, public safety, public networks, auto-vehicles, and robots in smart cities. The statistics indicate that recently there are great interests of applying CL methodologies to various areas of smart cities. Also, at the end of this section, we list numerous public datasets used in smart city research, as in Table \ref{tab:datasets}.

\begin{table*}[htbp]
\centering
\caption{Application domain categories of CL methods in smart city and their methods categories}
\label{tab:application}
\begin{tabular}{@{}lllccc@{}}
\toprule
\textbf{Domain} & \textbf{Subdomain}  & \textbf{Methods} & \textbf{Regularization-based} & \textbf{Replay-based} & \textbf{Architecture-based}
\\ \midrule
\multirow{19}{*}{\textbf{Transportation}}
    & \multirow{8}{*}{Traffic Flow Prediction} 
    & R2C-SVR-IL\cite{xiaoShorttermTrafficVolume2019} & \checkmark \\
    & & STMP\cite{nallaperumaOnlineIncrementalMachine2019} & \checkmark \\ 
    & & TF-Net\cite{yuIncrementalLearningBased2021} & \checkmark \\
    & & IL-TFNet\cite{shaoTrafficFlowPrediction2021} & \checkmark \\
    & & FpC\cite{lanzaUrbanTrafficForecasting2023} & \checkmark \\
    & & TrafficStream\cite{chenTrafficStreamStreamingTraffic2021} & \checkmark & \checkmark \\ 
    & & STKEC\cite{wangKnowledgeExpansionConsolidation2023} & \checkmark & \checkmark \\ 
    & & iETA \cite{hanIETARobustScalable2023} & \checkmark & \checkmark \\ 
    \cmidrule(lr){2-6}
    & \multirow{4}{*}{Traffic Video Analysis}
    & ArcVideo\cite{zhangDataEfficientContinuousLearning2022}  &  & \checkmark \\ 
    & & PAVEMENT\cite{maipraditPAVEMENTPassingVehicle2022}  & \checkmark \\
    & & FLCB\cite{gaoForgetLessCount2023} & \checkmark \\
    & & FedSTIL\cite{zhangSpatialTemporalFederatedLearning2023} &  & \checkmark \\ 
    \cmidrule(lr){2-6}
    & \multirow{7}{*}{Traffic Trajectory Analysis }
    & IKASL\cite{bandaragodaTrajectoryClusteringRoad2019} & & & \checkmark \\ 
    & & GWR\cite{tenzerLearningCitywidePatterns2022}  & & & \checkmark \\ 
    & & GCMTP\cite{maContinualMultiAgentInteraction2021} &  & \checkmark \\ 
    & & D-GSM\cite{linContinualInteractiveBehavior2023} &  & \checkmark \\ 
    & & CLTP-MAN\cite{yangContinualLearningbasedTrajectory2022} &  & \checkmark \\ 
    & & CL-SGR\cite{wuContinualPedestrianTrajectory2023} &  & \checkmark \\ 
    & & SCL-PP\cite{knoedlerImprovingPedestrianPrediction2022}  & \checkmark & \checkmark & \\ 
    \midrule
\multirow{6}{*}{\textbf{Environment}}
    & \multirow{2}{*}{Air Quality Prediction}
    & CEDUP\cite{wuCEDUPUsingIncremental2023} &  & \checkmark \\ 
    & & WISDOM\cite{xuSpatioTemporalMultiTaskLearning2021} & \checkmark \\
    \cmidrule(lr){2-6}
    & \multirow{1}{*}{Pollution Classification}
    & IEL\cite{hanPrototypeEnhancementBasedIncremental2023} & \checkmark \\
    \cmidrule(lr){2-6}
    & \multirow{3}{*}{Remote Sensing}
    & SARDINE\cite{dasSARDINESelfAdaptiveRecurrent2020}   & & & \checkmark \\ 
    & & LIL\cite{luLILLightweightIncremental2022}  & & & \checkmark \\ 
    & & DILRS\cite{ruiDILRSDomainIncrementalLearning2023}  & \checkmark & & \checkmark \\
    \midrule
\multirow{3}{*}{\textbf{Public Health}}& \multirow{3}{*}{/} 
    & MCLNP\cite{wangUncertaintyEstimationNeural2022} & \checkmark \\
    & & ANNAIL\cite{farooqNovelAdaptiveDeep2020} & \checkmark \\
    & & SEIRD-IL\cite{camargoIncrementalLearningApproach2022}  &  & \checkmark \\ 
    \midrule
\multirow{9}{*}{\textbf{Public Safety}}
    & \multirow{6}{*}{Human Activity Recognition}
    & FIRF\cite{huNovelFeatureIncremental2019} & \checkmark \\
    & & CITS\cite{xueliClassIncrementalTemporalSpatial2020} &  & \checkmark \\ 
    & & HAR-IL\cite{younanDeepIncrementalLearning2022} &  & \checkmark \\ 
    & & Solver-CM\cite{guptaContinualLearningMultivariate2021} &  & \checkmark \\ 
    & & DIM-BLS\cite{xiaoBaselineModelTraining2021}  & & \checkmark & \checkmark \\ 
    & & TTD\cite{yinContinualLearningAttentive2023}  & \checkmark \\ 
    \cmidrule(lr){2-6}
    & \multirow{1}{*}{Natural Disasters Detection}
    & DENet\cite{wangDomainincrementalLearningFire2023}  & \checkmark \\ 
    \cmidrule(lr){2-6}
    & \multirow{2}{*}{Power System Analysis} 
    & Dynamic-IL\cite{veerakumarDynamicIncrementalLearning2023}  &  & \checkmark \\ 
    & & CSTWPP\cite{huVeryShorttermSpatial2020} & \checkmark \\
    \midrule
\multirow{19}{*}{\textbf{Public Networks}}
    & \multirow{3}{*}{Internet Traffic Classification} 
    & ILAs\cite{eldhaiTrafficClassificationBased2022} & \checkmark \\
    & & ISVM\cite{sunInternetTrafficClassification2018} &  & \checkmark \\ 
    & & iCarl+\cite{bovenziFirstLookClass2021}  & \checkmark & \checkmark & \\ 
    \cmidrule(lr){2-6}
    & \multirow{3}{*}{Social Event Representation} 
    & HyperHawkes\cite{dubeyContinualLearningTimetoEvent2022} & \checkmark \\
    & & oHIML\cite{zhaoSpatioTemporalEventForecasting2021} & \checkmark \\
    & & DR-EMR\cite{vijayaraghavanLifelongKnowledgeEnrichedSocial2021}  & & & \checkmark \\ 
    \cmidrule(lr){2-6}
    & \multirow{5}{*}{Social Event Detection} 
    & KCN\cite{caoIncrementalEventDetection2020}  & \checkmark \\
    & & KPGNN\cite{caoKnowledgePreservingIncrementalSocial2021}  & & & \checkmark \\ 
    & & FinEvent\cite{pengReinforcedIncrementalCrossLingual2023}  & & & \checkmark \\ 
    & & LL-ED\cite{yuLifelongEventDetection2021}  & \checkmark & \checkmark & \\ 
    & & EMP\cite{liuIncrementalPromptingEpisodic2022}  & \checkmark & \checkmark & \\ 
    \cmidrule(lr){2-6}
    & \multirow{8}{*}{Recommendation} 
    & GraphSAIL\cite{xuGraphSAILGraphStructure2020} & \checkmark \\
    & & SACDLD\cite{wangGraphStructureAware2021} & \checkmark \\
    & & GDumb\cite{ahrabianStructureAwareExperience2021} &  & \checkmark \\ 
    & & FIRE\cite{xiaFIREFastIncremental2022} &  & \checkmark \\ 
    & & FMHP\cite{cuiEventbasedIncrementalRecommendation2023} &  & \checkmark \\ 
    & & STS-Rec\cite{amiratIncrementalTreebasedSuccessive2023} & & & \checkmark \\ 
    & & DEGC\cite{heDynamicallyExpandableGraph2023} & & & \checkmark \\ 
    & & IGC\cite{dingCausalIncrementalGraph2022}  & \checkmark \\ 
    \midrule

\multirow{5}{*}{\textbf{Robots}}& \multirow{5}{*}{/} 
    & AirLoop\cite{gaoAirLoopLifelongLoop2022} & \checkmark \\
    & & ISTMM\cite{kanazawaIncrementalLearningSpatialTemporal2019} & \checkmark \\
    & & CAA-HRI\cite{maharjanContinualLearningAdaptive2022} & \checkmark \\
    & & DVS-IL\cite{lunguFastEventdrivenIncremental2019}  & \checkmark & \checkmark & \\ 
    & & TCBLS\cite{caoIncrementalLearningRemaining2023}  & & & \checkmark \\ 
    
    \bottomrule
\end{tabular}
\end{table*}

\subsection{Transportation}
The transportation systems of modern cities can be extremely complex. Physically, there are now very rich commuting options for residents to choose from, such as bus, subway, taxi, biking, or walking;  technologically, many digital devices such as GPS, sensors, and cameras are deployed everywhere. Hence, perceiving, recording, and managing rich and changing transportation information becomes a non-trivial challenge. So people are leveraging continual learning to deal with (near) real-time traffic changes and incidents. Below we mainly introduce three sub-areas: traffic flow prediction, traffic video analysis, and trajectory analysis.

\subsubsection{Traffic flow prediction}
\begin{figure}[!h]
\centering
\subfloat[Station traffic flow]{
\includegraphics[width=2in]{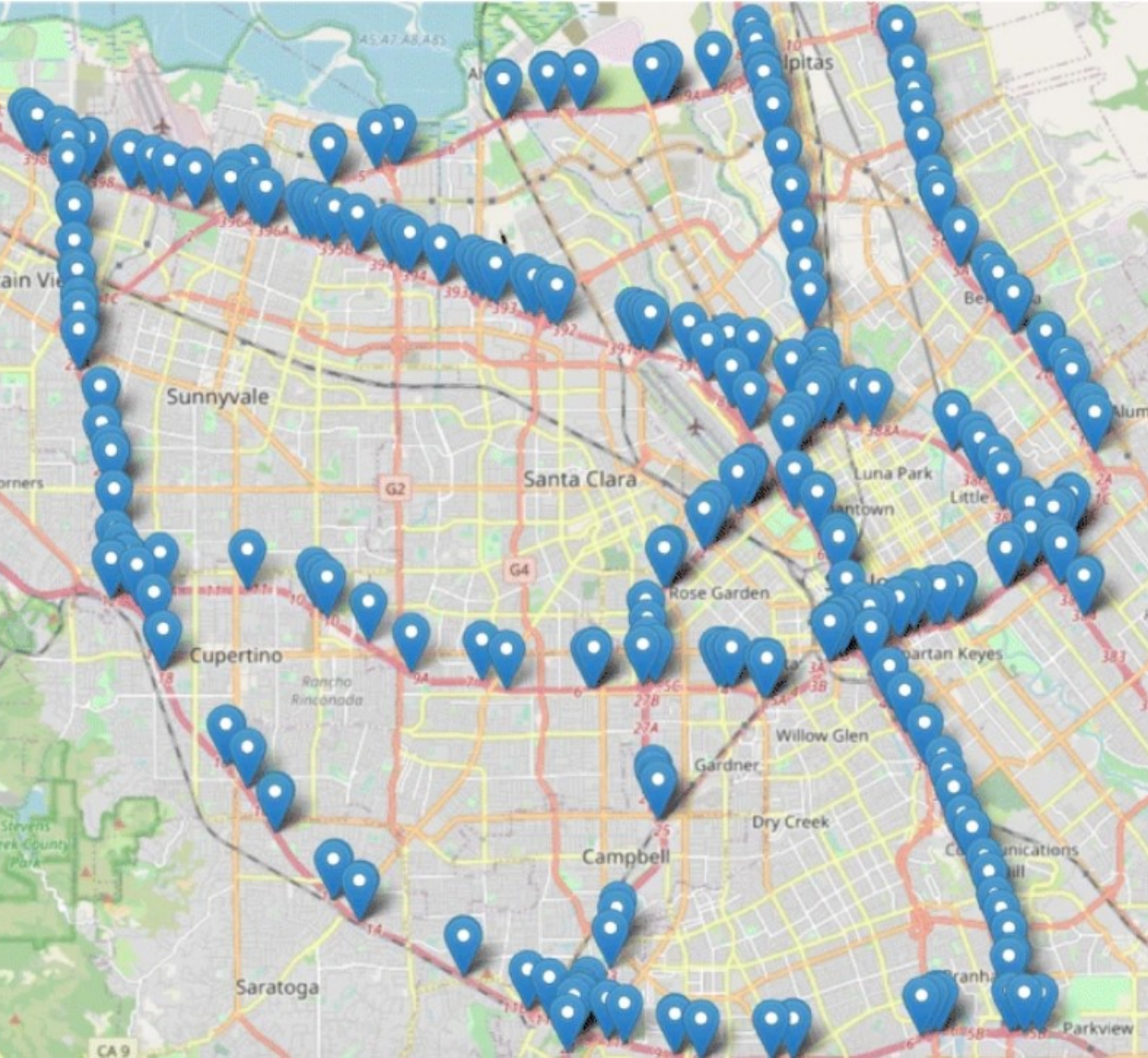} \label{fig:traffic_flow:pems}
}
\subfloat[Region traffic flow]{
\includegraphics[width=1.4in]{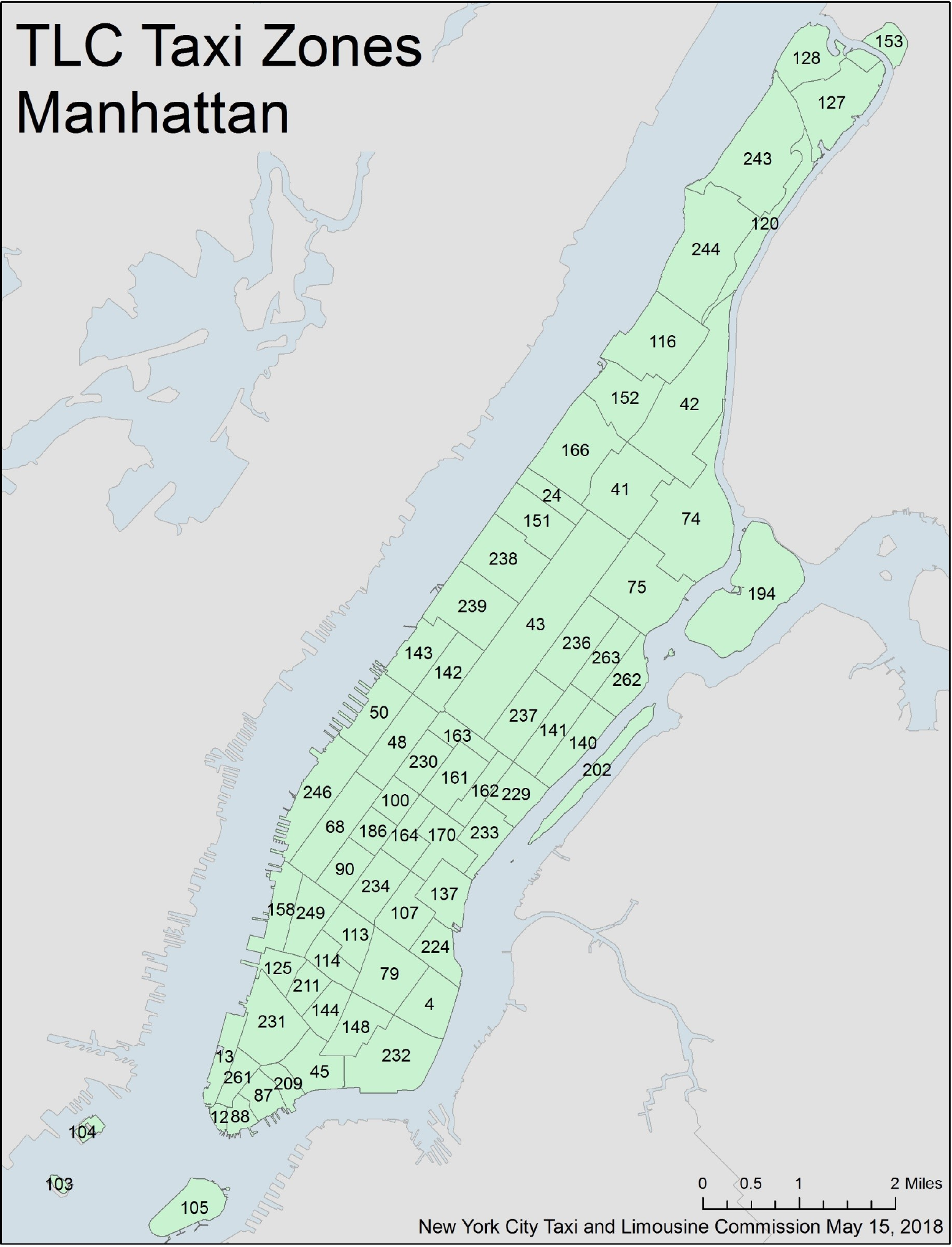} \label{fig:traffic_flow:nyc}
}
\caption{Two different types of data recording for traffic flow}
\label{fig:traffic_flow}
\end{figure}

Prediction of traffic flow on urban roads is one important task in intelligent transportation systems (ITSs). Traffic flow can be denoted as 1) the speed and counts of vehicles crossing some sections or 2) traffic inflow and outflow in certain regions, as shown in Fig.\ref{fig:traffic_flow}. Usually, both forms above are modeled as graph structures $\mathcal{G}$ and share the same prediction goals defined as follows.
\begin{equation}
    \label{eqn-traffic-flow}
    \{ X_{t-M+1},  \cdots , X_{t};\mathcal{G} \}\overset{f_\theta}{\rightarrow} \{ X_{t+1},  \cdots , X_{t+N}  \} 
\end{equation}
where the $X_t$ denotes the features of the traffic flow at a timestamp $t$, and $f_\theta$ denotes the prediction model.

There is a long history and many well-studied models on urban traffic flow prediction, from traditional time series methods, such as Auto-Regressive Integrated Moving Average (ARIMA) \cite{JSSv027i03}, Support Vector Regression (SVR) \cite{smolaTutorialSupportVector2004} and Gradient Boosting Decision Tree (GBDT) \cite{natekinGradientBoostingMachines2013}, to deep learning methods \cite{zhangDNNbasedPredictionModel2016, wuGraphWavenetDeep2019, yuSpatiotemporalGraphConvolutional2018}. Such models are mainly devoted to solving static data and thus become very limited in dealing with dynamic transportation networks and live flows. 

Introducing continual learning into traffic flow prediction can effectively mitigate the above problems. 
In early works, CL was used to solve the concept drift problem in non-stationary data~\cite{xiaoShorttermTrafficVolume2019,nallaperumaOnlineIncrementalMachine2019}. 
Later, some other works focused on meeting the real-time requirements of traffic flow prediction and reducing training costs through CL~\cite{yuIncrementalLearningBased2021, shaoTrafficFlowPrediction2021, lanzaUrbanTrafficForecasting2023}. 
More recent works have combined graph neural networks (GNNs) and CL to perform traffic flow prediction. These works centered more on solving the catastrophic forgetting problem. We mention some representative works below.

TrafficStream~\cite{chenTrafficStreamStreamingTraffic2021} first utilized CL to solve the problems of traffic network expansion and traffic flow evolution in a long-term streaming network. The network is defined as a sequence of evolving snapshots: $ \mathcal{G} = (\mathcal{G}_1, \mathcal{G}_2, \cdots, \mathcal{G}_\mathcal{T})$. The goal of TrafficStream is to learn a series of functions $\Psi = (\Psi_1, \Psi_2, \cdots, \Psi_\mathcal{T})$ to predict traffic flow series:
\begin{equation}
    \label{eqn-trafficStream-goal}
    \Psi_i^* = \arg\min_{\Psi_i} \left \| \Psi_\mathcal{T}(X_{\mathcal{T}-}) - X_{\mathcal{T}+}  \right \| ^2
\end{equation}
where the $X_{\mathcal{T}-}$ and $X_{\mathcal{T}+}$ denote the past and the future (ground-truth) traffic flow data, respectively. 

STKEC~\cite{wangKnowledgeExpansionConsolidation2023} also dealt with constantly evolving and expanding traffic networks. Specifically, incremental nodes and edges representing new sensor stations are continuously added to the topological graph over time. STKEC follows similar problem settings and notion definitions as TrafficStream.
iETA \cite{hanIETARobustScalable2023} is another work studying evolving traffic conditions, yet not on expanding traffic networks. They were all motivated by the fact that periodically retraining the prediction model is expensive while updating the predictors based on incremental data is more realistic.

\subsubsection{Traffic video analysis}

Traffic video analysis can be used for traffic monitoring, traffic analysis, traffic scheduling, and other applications~\cite{olatunjiVideoAnalyticsVisual2019, huEdgeBasedVideoAnalytics2023}. 
We categorize a few tasks related to traffic video analysis, including video classification~\cite{tranVideoClassificationChannelSeparated2019}, object detection~\cite{zouObjectDetection202023}, and semantic segmentation~\cite{kunduFeatureSpaceOptimization2016}.
Usually, the above works primarily focus on static tasks so have not addressed dynamically evolving tasks. For instance, the model needs to recognize vehicles and pedestrians under different weather conditions, as well as newly introduced trucks and trains on the road. Below we introduce some CL-related works.

VPaaS~\cite{zhangServerlessCloudFogPlatform2021} classified \textbf{Video Analysis Systems} into two categories, \textit{client-driven} and \textit{cloud-driven}. VPaaS posed three challenges regarding cloud-driven methods, bandwidth and latency in cloud transmission, data drift in fixed and pre-trained models, and inherent pipeline problems.
In contrast, Ekya~\cite{romilbhardwajEkyaContinuousLearning2022} focused on client-driven methods, also known as edge computing. To overcome the data drift problem on edge devices, Ekya utilizes CL techniques to maintain the performance of compressed models across many tasks. 
Besides the data drift problem, ArcVideo \cite{zhangDataEfficientContinuousLearning2022} further considered the human labeling cost and edge storage cost incurred by using CL methods. 
Specifically, new video frames and manual labels are required constantly, which increases human labor. Meanwhile, the replay-based approach consumes a lot of memory space, which is unfriendly for the storage in edge devices. C-EC \cite{nanLargescaleVideoAnalytics2023} leveraged the advantages of both client-driven and cloud-driven methods, helping edge models learn from real-time video and prevent forgetting caused by data drift.
Based on visual analysis tasks, PAVEMENT \cite{maipraditPAVEMENTPassingVehicle2022} proposed to compensate for the environmental impact of existing vehicle detection methods by integrating non-video sensors such as vibration sensors and Doppler sensors. Meanwhile, a CL technique is used to reduce the cost of human labeling and calculation overhead in the training process. 


In addition to vehicle detection, pedestrian detection is also an important application of traffic video analysis. \textbf{Crowd-counting} can be used to predict the number of people in images or videos and help manage crowded scenes in real-time. FLCB \cite{gaoForgetLessCount2023} proposed a lifelong crowd-counting task to obtain an optimal crowd-counting function on a series of $\mathcal{T} $ domain datasets. These different domains usually come from different locations of cameras, such as on a street, in a park, or in a gym. FLCB attempted to achieve a globally optimal performance on all domains through the CL technology, rather than a specific domain.

\textbf{Person re-identification} (ReID) is another application in pedestrian detection. Person ReID aims to identify the same person from cameras with different perspectives. Researchers first noticed that classic methods have many limitations in the real-world environment. For example, ReID data are constantly acquired from new locations or domains. 
GwFReID \cite{wuGeneralisingForgettingLifelong2021} believed there are three major challenges in the lifelong ReID task - that are zero-shot problems, incremental domains and classes, and imbalance classes. Meanwhile, AKA \cite{puLifelongPersonReIdentification2021} proposed that the Lifelong ReID needs to have the generalization ability to the unseen classes, and the classification ability to the fine-grained inter-class appearance variations. Subsequently, PTKP \cite{geLifelongPersonReidentification2022} took the fast domain adaptation ability as the goal of lifelong person ReID. PTKP pointed out that GwFReID and AKA do not consider the issues of task-wise domain gap, so their models can not learn task-shared knowledge very well. Following the above existing works, MEGE \cite{puMemorizingGeneralizingFramework2023} further studied the generalized representation of lifelong ReID models without forgetting the knowledge, and considered underlying adjacent relations between samples. Meanwhile, KRC \cite{yuLifelongPersonReidentification2023} regarded the lifelong ReID as a fine-grained open-set problem. KRC believed that existing lifelong person ReID works are mainly focused on preventing the forgetting problem. However, it is also important that the model can have positive FWT and BWT by the CL technology.
Based on the above person ReID works, FedSTIL \cite{zhangSpatialTemporalFederatedLearning2023} further considered privacy and security issues during data transmission and model training, thus a federated learning technique was used to make the edge clients obtain lifelong learning ability. In the setting of Federated Lifelong Person ReID, each edge client $c$ learns from streaming datasets $\mathcal{D}_{\tau}^{c}$, which denotes the $\tau$-th incremental datasets at edge client $c$. The goal of FedSTIL was to achieve lifelong learning for a single client and joint learning across clients while avoiding the sharing of sensitive information among edge clients and the center cloud.

\subsubsection{Trajectory analysis}
While lots of traffic analysis concentrates on traffic flow or congestion, there are also research works on comprehensive traffic condition analysis, a domain we refer to as \textbf{Trajectory Monitoring}.
Trajectory data can be collected from GPS, cell towers, and Wi-Fi to help city managers, drivers, or pedestrians know better about the traffic conditions from both citywide and individual perspectives.
By surveying the previous works, we categorize the applications as trajectory prediction \cite{liuSocialGraphTransformer2022, luVehicleTrajectoryPrediction2023}, trajectory clustering \cite{wangVehicleTrajectoryClustering2021, bianSurveyTrajectoryClustering2018}, and Path Planning.

IKASL \cite{bandaragodaTrajectoryClusteringRoad2019} proposed an incremental trajectory cluster algorithm to represent hyper-dimensional trajectories and to capture the time variation of traffic trajectories. IKASL segments and profiles raw road traffic data into distinct trajectory clusters, continually refining these profiles over time to adapt to evolving traffic patterns. This approach offers a clear understanding of traffic flow dynamics and serves as a foundation for effective traffic management strategies.
GWR~\cite{tenzerLearningCitywidePatterns2022} studied the mobility of human populations based on the trajectory datasets. In this work, the authors focused on resolving the forgetting problem caused by the concept shift of streaming data and further implemented a detection algorithm of the concept shift and assessment of their harmful effects.

\begin{figure}[!h]
\centering
\includegraphics[width=3.4in]{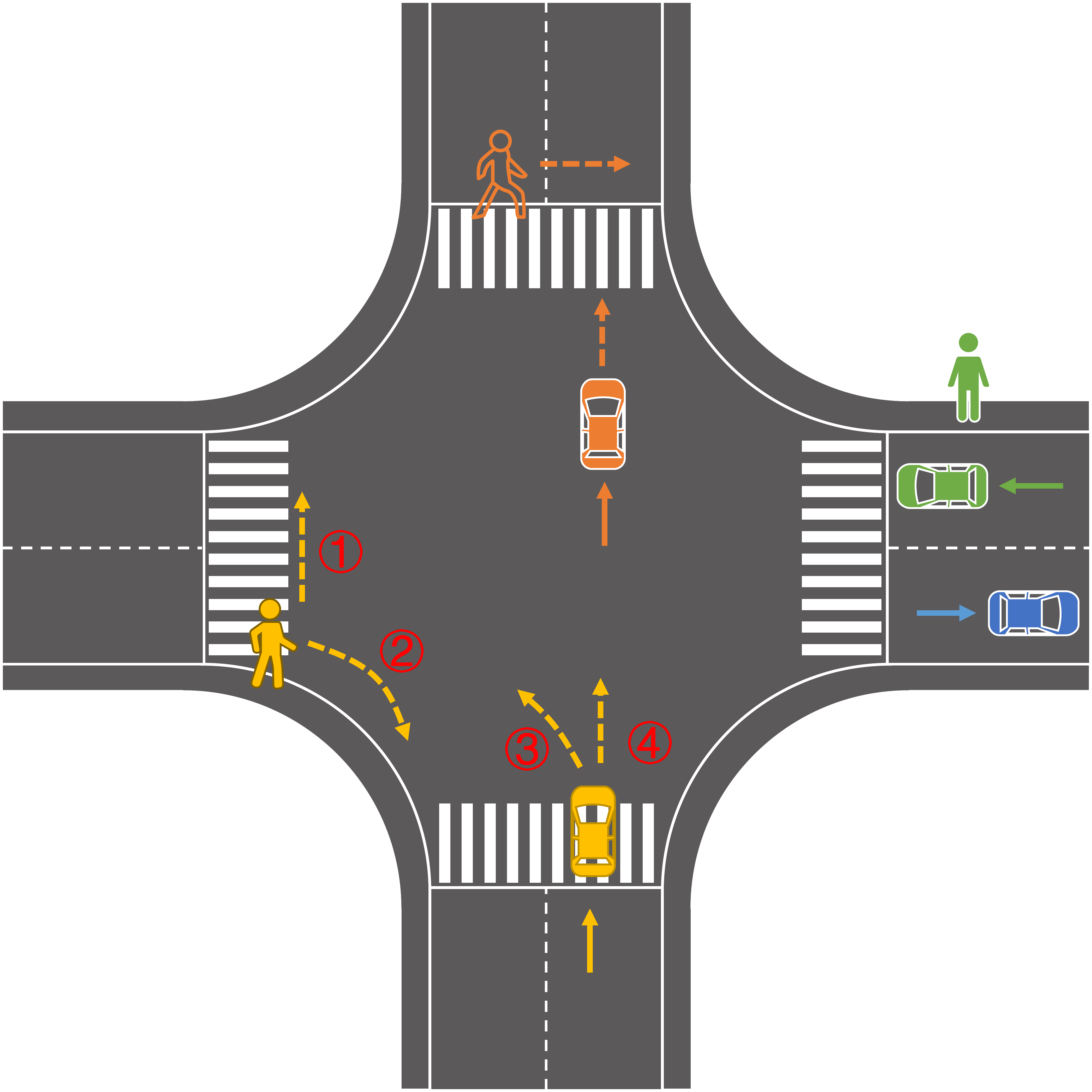}
\caption{A crossroads scenario for trajectory prediction. The solid line represents the agent's historical trajectory, and the dashed line represents the predicted trajectory. We observe a green vehicle and pedestrian stationary, while an orange vehicle and pedestrian are in motion. Their trajectories exhibit a degree of predictability. Conversely, the trajectory of the yellow vehicle and pedestrian has just started, introducing significant uncertainty. The objective of trajectory prediction is to anticipate future behavior within such intricate scenarios precisely.}
\label{fig:traj-pred}
\end{figure}

In addition to analyzing traffic trajectories at a citywide scale, there has been more attention on analyzing trajectories from an individual perspective, such as agents like vehicles and pedestrians. \textbf{Trajectory Prediction} is a prominent area of interest within this research domain. The trajectory prediction task involves forecasting the future paths of one or more agents using historical trajectory data and current road conditions. 
Accurate trajectory prediction is crucial for enhancing the safety of autonomous driving systems, enabling autonomous vehicles (AVs) to anticipate and avoid pedestrians and other vehicles effectively \cite{rudenkoHumanMotionTrajectory2020, huangSurveyTrajectoryPredictionMethods2022}. Fig. \ref{fig:traj-pred} shows a trajectory prediction task under complex road conditions. 

SILA \cite{habibiSILAIncrementalLearning2020} first raised the problem that the offline setting and batch learning setting limit the ability of pedestrian trajectory prediction models in terms of responding to environment changes and their generalization ability. Therefore, through the CL technology, the models are allowed to update the flexibly while adding the incremental few-shot data. Meanwhile, the features not exposed to historical data also enhance privacy. In SILA, an algorithm was designed to learn and update motion primitives and transitions incrementally.
SCL \cite{knoedlerImprovingPedestrianPrediction2022} aimed to help autonomous mobile robots realize the accurate prediction of the trajectories from pedestrians around the robots. Moreover, SCL proposed a self-supervised continual learning framework to achieve an online trajectory prediction.
CLTP-MAN \cite{yangContinualLearningbasedTrajectory2022} and CL-SGR \cite{wuContinualPedestrianTrajectory2023} were the following studies that further explore different CL techniques to solve the CF problem in pedestrian trajectory predictions.
GCMTP \cite{maContinualMultiAgentInteraction2021} was the earliest study on multi-agent trajectory prediction in the CL setting. Compared to single-agent trajectory prediction, multi-agent trajectory prediction coordinates multiple agents, usually referring to vehicles, simultaneously. So that their predicted trajectories do not collide with each other. GCMTP believed that the forgetting problem in the trajectory prediction task would take the form of the location changing. That is, multi-agent behaviors from a new interaction location may be very different from old interactions, so the model may prefer to learn the trajectory pattern from the current location and forget the historical locations.
IPCC-TP \cite{zhuIPCCTPUtilizingIncremental2023} proposed to utilize the Incremental Pearson Correlation Coefficient to obtain an optimal multi-agent trajectory prediction.
GRTP \cite{baoLifelongVehicleTrajectory2023} designed a Lifelong Vehicle Trajectory Prediction Framework for reliable autonomous driving, hoping the model can maintain consistent performance under different traffic circumstances.
Similarly, D-GSM \cite{linContinualInteractiveBehavior2023} was proposed to predict vehicle trajectory for autonomous driving in the CL scenarios.

\subsection{Environment}
Environment-friendly is critical to modern urban development, and technologies should constantly improve upon this target. To achieve long-term environmental monitoring and protection, continual learning, because of its cost-effectiveness in nature, is gradually being used in various environmental applications in smart cities. Below we introduce air quality control, pollution control, and remote sensing.

\subsubsection{Air quality prediction}
With urbanization and industrialization, air quality has become one critical environmental problem in cities. To monitor and protect air quality, many deep learning methods \cite{maoModelingAirQuality2021, yiDeepDistributedFusion2018} have been proposed to achieve the air quality prediction.
To model municipal-level carbon emissions, CEDUP \cite{wuCEDUPUsingIncremental2023} explored carbon emission distribution through incremental learning modeling, which can prevent provincial-level patterns from being forgotten while learning about municipal-level patterns.
Climate prediction plays a crucial role in the environmental monitoring of smart cities. Based on the combination of CL and spatio-temporal prediction models, SMART \cite{xuSpatioTemporalMultiTaskLearning2021} designed a Weight Incremental Spatio-Temporal Multi-Task Learning Algorithm (WISDOM), an incremental learning algorithm to achieve incremental learning on temporal-spatial data and can easily adapt to the dynamic spatio-temporal domain. Furthermore, SMART attempted to address the limitation of traditional methods where models cannot incorporate known spatio-temporal knowledge. For example, when models are learning local climate patterns, they should also be able to use larger-scale weather patterns.

\subsubsection{Pollution classification}
Urban pollutants accumulate with the operation and production of cities. Pollution, such as urban garbage, air pollution, and water pollution, if not cleaned up in time, can hurt people's health and happiness. There have been a lot of DNN-based methods to predict and classify pollution \cite{attaallahSMOTEDNNNovelModel2022, maoRecyclingWasteClassification2021}.
Due to the increase of urban garbage, recycling of garbage resources has become a challenging problem. In a practical application environment, garbage groups will change over time, and it is difficult to ensure the accuracy of classification by using static models. This problem can be partially solved by CL. Han et al. \cite{hanPrototypeEnhancementBasedIncremental2023} propose the IEL algorithm, which can improve the garbage classification accuracy and generalization ability of incremental learning.

\subsubsection{Remote sensing}
Remote sensing has played an important role in urban planning, natural disaster monitoring, and safety prevention. Remote sensing images captured by artificial satellites can provide continuous weather observation, building observation, and ecological environment observation covering a large area of the city, which is very useful for detailed monitoring and environmental quality assessment.
However, with the accumulation of new data and the drift of data distribution, the traditional static model is difficult to deal with in a complex environment, so using CL is an important method to solve the above challenges. 

In recent years, some work has introduced CL into remote sensing. 
SARDINE \cite{dasSARDINESelfAdaptiveRecurrent2020} was an earlier work in this field, in which the proposed model's incremental learning capability provides additional advantages when addressing variations in spatial contexts across widely dispersed regions, as observed in satellite imagery. 
LIL \cite{luLILLightweightIncremental2022} proposed two new challenge issues of remote Sensing in CL, which are the task-sharing feature extractor large-capacity issue and task-specific module redundancy issue. 
DILRS \cite{ruiDILRSDomainIncrementalLearning2023} instead focused on the domain-incremental problem. Specifically, the domain shift in the remote sensing task includes inconsistent class distributions across various regions, along with disparities in spatial resolutions and spectral divergence of specific object categories across different sensors.
Unlike the classification task, the counting task is a regression problem. DMD \cite{wuDensityMapDistillation2023} discusses the application of CL for counting systems of urban objects such as buildings, vehicles, and ships.

\subsection{Public Health}
Citizens' health in city management comes in the first place. Take epidemics as an example, when a global pandemic such as COVID-19 threatens people's health, predicting uncertainty in the pandemic is a huge challenge. COVID-19 is characterized by rapid variation and infection, and traditional pre-training models are insufficient to handle the fast dynamic change. Hence, continual learning can be used here for its ability to adapt to the uncertainty caused by environmental changes, especially in non-stationary environments. 
Farooq et al. \cite{farooqNovelAdaptiveDeep2020} proposed a DNN-based and data stream-guided real-time online incremental learning algorithm (ANNAIL) to study the transmission dynamics and prevention mechanism for COVID-19. Their work can help forecast the pandemic and suggest relevant policies.  
For a similar motivation, SEIRD-IL \cite{camargoIncrementalLearningApproach2022} proposed an incremental learning approach for online predictions of epidemic diseases based on a dynamic ensemble method. And MCLNPs \cite{wangUncertaintyEstimationNeural2022} studied the uncertainty problem of a CL method in the non-stationary environment of the epidemic disease prediction task.

\subsection{Public Safety}
The operation of modern cities is supported by multiple infrastructures such as food, power, water, gas, and network supplies. Monitoring and detection of faults in these systems are necessary to guarantee public safety. In addition, cities may suffer from sudden disasters like accidents and natural disasters, and thus we also need to maintain real-time intelligent models that have learned the latest changes.

\subsubsection{Human activity recognition}
One important aspect of public safety is human activity recognition(HAR). HAR recognizes human activities through wireless sensor data and can help prevent extreme public incidents from happening. In the real world, the sensor or edge equipment faces challenges such as privacy protection, new behavior or new user recognition, and power efficiency. 
Recently, many works are applying the continual learning logic to address the above challenges. Some researchers applied CL to reduce the training time of edge equipment \cite{huNovelFeatureIncremental2019, younanDeepIncrementalLearning2022}, update the model without the historical privacy data \cite{guptaContinualLearningMultivariate2021}, identify new users and their new actions \cite{xueliClassIncrementalTemporalSpatial2020, xiaoBaselineModelTraining2021, guptaContinualLearningMultivariate2021}, or solve the sensor temporal data increment and catastrophic forgetting problems \cite{yinContinualLearningAttentive2023}.

\subsubsection{Natural disasters detection}
Natural disasters such as wildfires, earthquakes, and floods are devastating to people's lives and properties. Detecting these natural disasters in time can help dramatically reduce potential losses. Some researchers proposed to detect disasters by using social media or satellite imagery \cite{saidNaturalDisastersDetection2019, weberDetectingNaturalDisasters2020}.
To develop fire detection models based on sensor data, it is necessary to balance the real-time performance and computational efficiency of the model, and CL is a highly effective way to tackle such as problem. Wang et al. \cite{wangDomainincrementalLearningFire2023} proposed a domain-incremental fire detection method to enable incremental updates of models by continuously learning heterogeneous data.

\subsubsection{Power system analysis}
City supplying systems' stability has become a more important issue since the complexity of these systems increases, which in part leads to a higher rate of failures. Therefore, it is necessary to identify and classify the system faults. In a real system, a complete fault dataset cannot be obtained at once, which requires the model to have the ability to identify new faults. Veerakumar et al. \cite{veerakumarDynamicIncrementalLearning2023} introduced CL to power systems with the ability to address catastrophic forgetting.
CSTWPP \cite{huVeryShorttermSpatial2020} trained the model continually, enabling the model to serve online and solve the problems of processing non-stationary new data and consuming excessive computing resources for repeated training.

\subsection{Public Networks}
Many activities of modern cities take place in cyberspace. Therefore, both the local and global networks need to be securely monitored and managed. Here we discuss network traffic analysis, social events analysis, and recommendation system design from the view of continual learning.

\subsubsection{Internet traffic analysis}
The ability to identify flows and their related protocols in network traffic classification is necessary for many applications, such as security and quality of service (QoS). To enable traffic classification models to obtain large-scale data and real-time processing capabilities, CL is one of the important tools. Sun et al. \cite{sunInternetTrafficClassification2018} introduced an incremental SVMs (ISVM) model to reduce the high training cost of memory and CPU, such that the traffic classifier can achieve quick updates. Bovenzi et al. \cite{bovenziFirstLookClass2021} utilized iCarl to also solve the high-cost problem of retraining models. Eldhai et al. \cite{eldhaiTrafficClassificationBased2022} proposed four incremental learning algorithms to effectively identify concept drifts while using less memory and time.

\subsubsection{Social event analysis}
In the tasks of social event analysis, a realistic and challenging problem is to continually learn time-to-event models in an ever-changing environment while retaining previously learned knowledge. Dubey et al. \cite{dubeyContinualLearningTimetoEvent2022} and Vijayaraghavan \cite{vijayaraghavanLifelongKnowledgeEnrichedSocial2021} proposed continual learning-based representation learning approaches to address the above challenges. Zhao et al. \cite{zhaoSpatioTemporalEventForecasting2021} proposed an incremental multi-source feature learning algorithm to quickly learn the new missing patterns in real-time without retraining the whole model.
Social event detection also plays an important role in smart city management. There are already lots of works on how to continuously learn models for new event classes while not forgetting previously learned knowledge under the constraints of computational costs and storage budgets \cite{caoIncrementalEventDetection2020, caoKnowledgePreservingIncrementalSocial2021, yuLifelongEventDetection2021}. And more works followed in this line - EMP \cite{liuIncrementalPromptingEpisodic2022} introduced episodic memory prompts to explicitly retain the learned task-specific knowledge to relieve CF; FinEvent\cite{pengReinforcedIncrementalCrossLingual2023} proposed using an incremental learning framework to help the model continuously acquire, preserve, and extend the semantic space, and make use of its advantages to deal with the actual problems.

\subsubsection{Recommendation systems}
Recommendation systems have thrived in many parts of public networks. Retraining the model in practical applications is very time-consuming, and directly fine-tuning the model can also lead to catastrophic forgetting. To solve such problems, many studies have focused on recommendation systems. Also, graph neural networks are introduced as many user-item, user-user, and item-item relationships can be well represented by graphs. Therefore, recommendation systems based on continual graph learning \cite{xuGraphSAILGraphStructure2020, ahrabianStructureAwareExperience2021, wangGraphStructureAware2021, dingCausalIncrementalGraph2022, xiaFIREFastIncremental2022, amiratIncrementalTreebasedSuccessive2023, cuiEventbasedIncrementalRecommendation2023, heDynamicallyExpandableGraph2023} have become an important research direction.

\subsection{Robots}
Although intelligent robots aren't widely used in smart cities, they indirectly contribute to city management, production, and our daily lives. CL can play a crucial role in robot perception, recognition, action, and other processes.
Simultaneous localization and mapping (SLAM) systems are one of the most important components of modern robots. AirLoop \cite{gaoAirLoopLifelongLoop2022} utilized CL to minimize forgetting when training loop closure detection models incrementally. For recognition, Lungu et al. \cite{lunguFastEventdrivenIncremental2019} proposed a hand symbol recognition system that can incrementally train on the platforms with limited resources and recognize new symbols without forgetting the old symbols. 
Besides, Human-robot interaction is another big topic. Kanazawa et al. \cite{kanazawaIncrementalLearningSpatialTemporal2019} introduced a collaborative robot that can update the model adaptively. Social robots need a model with the ability to deal with individual differences and emotional changes in the process of interaction with people and to learn new emotions during the update of models \cite{churamaniContinualLearningAffective2020, irfanLifelongLearningPersonalization2022, maharjanContinualLearningAdaptive2022,castriContinualLearningCausal2023}.
In addition, there are a large number of specialized equipment in smart cities. Accurately predicting the remaining useful life of such equipment can increase the production efficiency. TCBLS \cite{caoIncrementalLearningRemaining2023} introduced CL in an online learning scenario where new data are constantly acquired. Faced with situations where newly acquired data and prediction accuracy are inadequate, online machine learning of new data and nodes can be implemented to adaptively update and upgrade the network.

\begin{table*}[htbp]
\centering
\caption{Open datasets for continual learning in smart cities.}
\label{tab:datasets}
\begin{tabular}{@{}lll@{}}
\toprule
\textbf{Domain} & \textbf{Datasets}  & \textbf{Links} \\ \midrule
\multirow{20}{*}{\textbf{Transportation}}
    & PEMS3-Stream &  https://github.com/AprLie/TrafficStream  \\
    & VicRoads & https://vicroadsopendata-vicroadsmaps.opendata.arcgis.com  \\ 
    & PeMS & http://pems.dot.ca.gov  \\ 
    & Madrid & https://datos.madrid.es/portal/site/egob \\ 
    & Barcelona & https://opendata-ajuntament.barcelona.cat/data/en/dataset/itineraris \\
    & Gdansk & https://doi.org/10.34808/8xkq-7714 \\
    & Turin & https://doi.org/10.5194/isprs-annals-IV-4-W7-3-2018 \\
    & IARAI & https://proceedings.mlr.press/v133/kopp21a.html \\
    & UTD19 & https://utd19.ethz.ch/ \\
    & DashCam & https://arxiv.org/pdf/2102.03012.pdf \\ 
    & Porto & https://ieeexplore.ieee.org/abstract/document/6532415 \\ 
    & ShanghaiTech & https://doi.org/10.1109/CVPR.2016.70 \\ 
    & UCF-QNRF & https://doi.org/10.1007/978-3-030-01216-8\_33 \\
    & NWPU-Crowd & https://doi.org/10.1109/TPAMI.2020.3013269 \\ 
    & JHU-Crowd++ & https://doi.org/10.1109/ICCV.2019.00131 \\
    & Market-1501 & https://doi.org/10.1109/ICCV.2015.133 \\
    & PKU-ReID & https://arxiv.org/pdf/1605.02464.pdf \\  
    & PersonX & https://arxiv.org/abs/1812.02162 \\ 
    & Prid2011 & http://dx.doi.org/10.1007/978-3-642-21227-7\_9 \\ 
    & DukeMTMC-reID & https://doi.org/10.48550/arXiv.1609.01775 \\ 
    \midrule
\multirow{3}{*}{\textbf{Environment}}
    & FASDD &  https://doi.org/10.57760/sciencedb.j00104.00103  \\
    & Huawei Cloud &  https://ieeexplore.ieee.org/abstract/document/9435085 \\
    & SMART-data &  https://github.com/Jianpeng-Xu/TKDE-SMART \\ 
    \midrule
\multirow{1}{*}{\textbf{Public Health}}
    & COVID & https://www.kaggle.com/fireballbyedimyrnmom/us-counties-covid-19dataset \\
    \midrule
\multirow{3}{*}{\textbf{Public Safety } }
    & WISDM & https://archive.ics.uci.edu/ml/machine-learning-databases/00507  \\ 
    & Anguita dataset &  https://sensor.informatik.uni-mannheim.de/\#dataset\_realworld \\ 
    & HAPT & https://www.esann.org/sites/default/files/proceedings/legacy/es2013-84.pdf \\   
    \midrule
\multirow{10}{*}{\textbf{Public Networks}}
    & MIRAGE-2019 &  http://traffic.comics.unina.it/mirage/app list.html \\ 
    & TOR dataset &  https://www.scitepress.org/PublishedPapers/2017/61056/61056.pdf \\ 
    & Lifelong Social Events Dataset & https://pralav.github.io/lifelong eventrep?c=10  \\ 
    & Yelp & https://www.kaggle.com/datasets/yelp-dataset/ \\
    & Meme & https://snap.stanford.edu/data/memetracker9.html \\ 
    & FewEvent & https://arxiv.org/pdf/1910.11621 \\
    & MAVEN & https://arxiv.org/pdf/2004.13590 \\
    & ACE05-EN &  https://www.ldc.upenn.edu/sites/www.ldc.upenn.edu/files/lrec2004-ace-program.pdf\\
    & Taobao2014 &  https://tianchi.aliyun.com/dataset/dataDetail?dataId=46\\ 
    & Nefix & https://academictorrents.com/details/9b13183dc4d60676b773c9e2cd6de5e5542cee9a \\ 
    \midrule
\multirow{5}{*}{\textbf{Auto-vehicle}}
    & ETH   & http://vision.cse.psu.edu/courses/Tracking/vlpr12/PellegriniNeverWalkAlone.pdf\\ 
    & PUCY & https://onlinelibrary.wiley.com/doi/10.1111/j.1467-8659.2007.01089.x \\ 
    & inD &  https://arxiv.org/pdf/1911.07602 \\ 
    & INTERACTION &  https://arxiv.org/pdf/1910.03088.pdf; \\ 
    & SDD &  https://infoscience.epfl.ch/record/230262/files/ECCV16social.pdf \\ 
    \midrule
\multirow{4}{*}{\textbf{Robots} }
    & TartanAir & https://arxiv.org/pdf/2003.14338  \\ 
    & Nordland & https://arxiv.org/pdf/1808.06516  \\ 
    & RobotCar &  https://doi.org/10.1177/0278364916679498 \\ 
    & AffectNet &  https://arxiv.org/pdf/1708.03985 \\ 
    \bottomrule

\end{tabular}
\end{table*}

\subsection{Open Datasets}
Lastly, we summarize a list of open datasets related to continual learning and smart city work in Table \ref{tab:datasets}. Some datasets are specially designed for CL tasks, which means the datasets are manually divided into several parts, as if they are sequential tasks. For example, PEMS3-Stream is built from PeMS by using seven years' data with each year being a task. In other research, authors group multiple datasets as one CL dataset. In this case, authors usually consider one as a training dataset and others as new sequential datasets for the latter can have significantly different distributions from the first one.

\section{Advanced Continual Learning Frameworks}
\label{sec-framework}

In this section, we introduce some of the latest continual learning frameworks combined with other machine learning paradigms. Such combinations are as expected since continual learning itself is sort of an \textquotedblleft add-on\textquotedblright\ process that is built on top of other learning tasks, which can vary a lot. For our purpose, we will review those advanced CL learning frameworks closely related to smart city research. They are continual graph learning, temporal continual learning, spatial-temporal continual learning, multi-modality continual learning, and federated continual learning.

\subsection{Continual Graph Learning}
Graphs are widely used for network data representation such as citation networks \cite{liuLinkPredictionPaper2019, luoPrototypeBasedInterpretabilityLegal2023}, social networks \cite{sunAligningDynamicSocial2023, liAdversarialLearningWeaklySupervised2019}, traffic networks \cite{zhaoTGCNTemporalGraph2020, baiAdaptiveGraphConvolutional2020, guoAttentionBasedSpatialTemporal2019}, and knowledge graphs \cite{chamiLowDimensionalHyperbolicKnowledge2020, zhouImprovingConversationalRecommender2020, xueKnowledgeGraphQuality2023}. In a graph, nodes represent entities and edges represent their relationships \cite{juComprehensiveSurveyDeep2023}. Analysis of graphs focuses on tasks such as node classification, link prediction, graph classification, and so on. Graph Neural Networks (GNNs) are a type of graph analysis method based on deep learning. With the steady progress of research, GNNs' focus is gradually changing from static graphs to dynamic ones. However, current research on dynamic graphs mainly focuses on changes on node features or edges, with less consideration on the addition of nodes and sub-graphs, which may cause catastrophic forgetting. 

\begin{figure}[!h]
\centering
\includegraphics[width=3.4in]{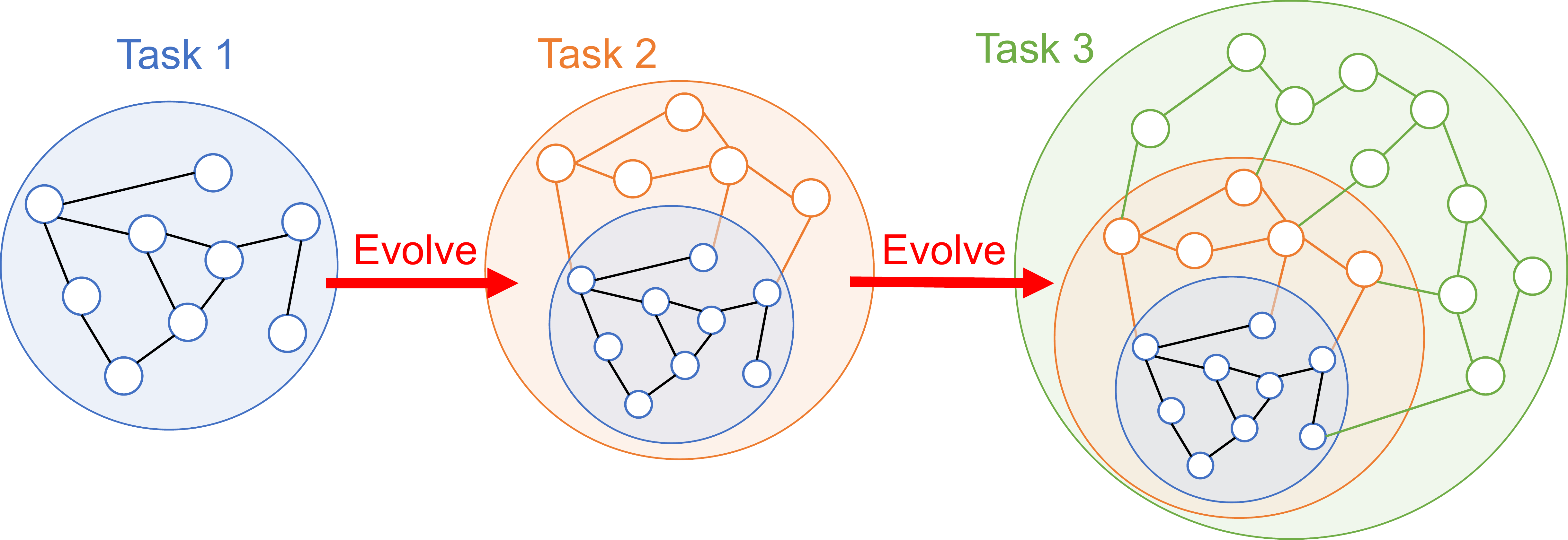}
\caption{Continual graph learning deals with the increment or decrement nodes and edges in a continuously evolving graph. The original graph as Task 1 is marked in blue. After two evolutions, the new graphs with incremental nodes and edges are marked in orange and green.}
\label{fig:GCL}
\end{figure}

Continual graph learning (CGL) studies how to implement CL methods on graph-structured data. While many CL methods are particularly designed for vision tasks, they cannot be directly applied to graph data, which are in a unique non-Euclidean data structure. The specificity of GCL is explained in Fig \ref{fig:GCL}. In general, GCL faces three challenges:
\begin{itemize}
    \item How to detect and represent new instances on a graph, including node-level and graph-level information, and keep accumulating new knowledge.
    \item How to achieve efficient graph learning without using the entire graph.
    \item How to avoid catastrophic forgetting and memorize what has been learned from previous graphs.
\end{itemize}

Currently, the majority of CGL works focus on the first challenge. For example, DyGNN \cite{maStreamingGraphNeural2020} proposed a general framework that can keep updating node information by capturing the sequential information of edges. At the same time, Han et al. \cite{hanGraphNeuralNetworks2020} focused on the same problem that how to deal with new or unseen data by using GNNs. Daruna et al. \cite{darunaContinualLearningKnowledge2021} attempted to represent new nodes and edges on the Knowledge Graph. Later, other works such as \cite{chenTrafficStreamStreamingTraffic2021} and \cite{wangKnowledgeExpansionConsolidation2023} focused on modeling graph expansions, specifically referring to the expansion of an urban road network. FILDNE \cite{bielakFILDNEFrameworkIncremental2021} studied how to apply static graph representation methods to dynamically incremental graphs. Recently, the unbalance between new and old classes on graphs is another research area, e.g. \cite{wangKnowledgeExpansionConsolidation2023}.

Streaming graph data is a common scenario in real-world graph learning applications, for example, to add or delete a node/edge in social networks with changing relationships. Wang et al. \cite{wangStreamingGraphNeural2020} were the early ones dealing with streaming graph data and overcoming forgetting by using data replaying and model regularization. In their following work, \cite{wangStreamingGraphNeural2022} proposed a streaming GNN by using the generative replay method. Other research covers more areas. Wang et al. \cite{wangIncrementalMobileUser2020} used knowledge graphs to model streaming events. DiCGRL \cite{kouDisentanglebasedContinualGraph2020} proposed a disentangle-based continual graph representation learning to alleviate the forgetting problem. IncreSTGL \cite{yuanIncrementalSpatioTemporalGraph2021} proposed an Incremental Spatio-Temporal Graph Learning framework that has an incremental graph representation learning module to refine and update query-POI interaction graphs in an online incremental fashion. FGN \cite{wangLifelongGraphLearning2022} converted a CGL problem to a regular graph learning problem so that GNN can inherit the lifelong learning techniques developed for convolutional neural networks.

Besides, some works focus on tackling catastrophic forgetting. To keep the long-term preference of users, GraphSAIL \cite{xuGraphSAILGraphStructure2020} implemented a graph structure preservation strategy that explicitly preserves each node’s local structure, global structure, and self-information.  
ER-GNN \cite{zhouOvercomingCatastrophicForgetting2021} stored knowledge from previous tasks as experiences and replayed them when learning new tasks to mitigate the catastrophic forgetting issue. Three experience node selection strategies, called mean of feature, coverage maximization, and influence maximization, were proposed to guide the process of selecting experience nodes. 
Another work using the replay-based method was \cite{galkeLifelongLearningGraph2021}. In this work, an incremental training method for lifelong learning on graphs was presented and a new measure based on k-neighborhood time differences to address variances in the historical data is proposed. Further, a Hierarchical Prototype Networks (HPNs)\cite{zhangHierarchicalPrototypeNetworks2023} was presented, which extracts different levels of abstract knowledge in the form of prototypes to represent the continuously expanded graphs. 

In addition, some works focus on the challenge regarding economic costs, like computational and memory consumption, which can be expensiveve especially in training deep models. IGCN \cite{xiaIncrementalGraphConvolutional2021} was proposed to learn new data and avoid the computational cost incurred when retraining GCN models. 
Ahrabian \cite{ahrabianStructureAwareExperience2021} presented a novel framework for incrementally training GNN models with an experience reply technique. After that, a novel Graph Structure Aware Contrastive Knowledge Distillation for Incremental Learning in recommence systems \cite{wangGraphStructureAware2021} was proposed to focus on the rich relational information in the recommendation context, and the contrastive distillation formulation was combined with intermediate layer distillation to inject layer-level supervision. More recently, a causal incremental graph convolution (IGC) \cite{dingCausalIncrementalGraph2022} approach was proposed, which consists of two new operators named IGC and colliding effect distillation (CED) to estimate the output of full graph convolution. 

Finally, other works consider CGL under more complex conditions, such as few-shot or multi-modality. The abovementioned tasks all assume sufficient labels, but in reality, there are many scenarios with insufficient labels of new nodes and edges, i.e., graph few-shot class-incremental learning (GFSCIL) problem. Geometer \cite{luGeometerGraphFewShot2022} learned and adjusted the attention-based prototypes based on the geometric relationships of proximity, uniformity, and separability of representations. A teacher-student knowledge distillation and biased sampling strategy were proposed to further mitigate the catastrophic forgetting and unbalanced labeling in GFSCIL. Under the same problem setting, a Hierarchical-Attention-based Graph Meta-learning framework (HAG-Meta) \cite{tanGraphFewshotClassincremental2022} presented a task-sensitive regularizer calculated from task-level attention and node class prototypes to mitigate overfitting onto either novel or base classes. To combine multiple modalities such as visual and textual features, the Multi-modal Structure-evolving Continual Graph Learning (MSCGL) model\cite{caiMultimodalContinualGraph2022} was proposed to simultaneously take social information and multi-modal information into account to build the multi-modal graphs. 
In addition, some important benchmark works were presented. Cart et al. \cite{cartaCatastrophicForgettingDeep2021} gave a benchmark for graph classification by experimenting in a robust and controlled framework. CGLB \cite{zhangCGLBBenchmarkTasks2022} systematically studied the task configurations in different application scenarios and developed a comprehensive continual graph learning benchmark curated from different public datasets.

\begin{table*}[htbp]
\centering
\caption{Advanced continual learning frameworks used in smart city research.}
\label{tab:framework}
\begin{tabular}{@{}lll@{}}
\toprule
\textbf{Learning Frameworks} & \textbf{Model}  & \textbf{Specific Task} \\ 
\midrule 
\multirow{14}{*}{\textbf{Continual Graph Learning}}
    & DyGNN\cite{maStreamingGraphNeural2020} & Computer Vision \\
    & FILDNE\cite{bielakFILDNEFrameworkIncremental2021} & Representation Learning \\ 
    & DiCGRL\cite{kouDisentanglebasedContinualGraph2020} & Representation Learning \\ 
    & IncreSTGL\cite{yuanIncrementalSpatioTemporalGraph2021} & Query-POI Matching \\ 
    & FGN\cite{wangLifelongGraphLearning2022} & Representation Learning \\ 
    & GraphSAIL\cite{xuGraphSAILGraphStructure2020} & Recommender Systems \\ 
    & ER-GNN\cite{zhouOvercomingCatastrophicForgetting2021} & Node Classification \\ 
    & HPNs\cite{zhangHierarchicalPrototypeNetworks2023} & Node Classification \\ 
    & IGCN\cite{xiaIncrementalGraphConvolutional2021} & Recommender Systems \\ 
    & IGC\cite{dingCausalIncrementalGraph2022} & Recommender Systems \\ 
    & Geometer\cite{luGeometerGraphFewShot2022} & Node Classification \\ 
    & HAG-Meta\cite{tanGraphFewshotClassincremental2022} & Node Classification \\ 
    & MSCGL\cite{caiMultimodalContinualGraph2022} & Multi-modal Node Classification \\ 
    & CGLB\cite{zhangCGLBBenchmarkTasks2022} & Benchmark \\ 
\midrule
\multirow{4}{*}{\textbf{Temporal Continual Learning}}
    & CLA\cite{philpsTemporalContinualLearning2020} & Finance Decision-making \\ 
    & Solver-CM\cite{guptaContinualLearningMultivariate2021} & Multivariate Time Series \\ 
    & HyperHawkes\cite{dubeyContinualLearningTimetoEvent2022} & Time-to-Event Modeling \\ 
    & TTD\cite{yinContinualLearningAttentive2023} & Temporal data classification \\ 
\midrule
\multirow{11}{*}{\textbf{Spatial-temporal Continual Learning}}
    & IL-GMM\cite{kanazawaIncrementalLearningSpatialTemporal2019} & Planning Motion \\ 
    & R2C\cite{xiaoShorttermTrafficVolume2019} & Traffic Volume Prediction \\ 
    & IKASL\cite{bandaragodaTrajectoryClusteringRoad2019} & Trajectory Clustering \\ 
    & CGM\cite{maContinualMultiAgentInteraction2021} & Trajectory Prediction \\ 
    & SARDINE\cite{dasSARDINESelfAdaptiveRecurrent2020} & Remote Sensing \\ 
    & CSTWPP \cite{huVeryShorttermSpatial2020} & Wind Power Forecasting \\ 
    & WISDOM\cite{xuSpatioTemporalMultiTaskLearning2021} & Multi-task Learning \\ 
    & TF-Net\cite{yuIncrementalLearningBased2021} & Traffic Flow Prediction \\ 
    & IL-TFNet\cite{shaoTrafficFlowPrediction2021} & Traffic Flow Prediction \\ 
    & TrafficStream\cite{chenTrafficStreamStreamingTraffic2021} & Traffic Flow Prediction \\ 
    & STKEC\cite{wangKnowledgeExpansionConsolidation2023} & Traffic Flow Prediction \\ 
\midrule
\multirow{7}{*}{\textbf{Multi-modality  Continual Learning}}
    & CLiMB\cite{srinivasanCLiMBContinualLearning2022} & Benchmark \\ 
    & MSCGL\cite{caiMultimodalContinualGraph2022} & Node Classification \\ 
    & CCMR\cite{wangContinualLearningCrossmodal2021} & Image-text Retrieval \\ 
    & BMU-MoCo\cite{gaoBMUMoCoBidirectionalMomentum2022} & Video-Language Modeling \\ 
    & VQACL\cite{zhangVQACLNovelVisual2023} & Visual Question Answering \\ 
    & Mod-X \cite{niContinualVisionLanguageRepresentation2023} & Vision-Language Representation \\ 
    & LMC\cite{chenContinualMultimodalKnowledge2023} & Knowledge Graph Construction \\ 
\midrule
\multirow{5}{*}{\textbf{Federated Continual Learning}}
    & FedWeIT\cite{yoonFederatedContinualLearning2021} & Image Classification \\ 
    & CFeD\cite{maContinualFederatedLearning2022} & Image Classification \\ 
    & FedSpace\cite{shenajAsynchronousFederatedContinual2023} & Image Classification \\ 
    & FedSTIL\cite{zhangSpatialTemporalFederatedLearning2023} & Person Re-identification   \\ 
    & FpC\cite{lanzaUrbanTrafficForecasting2023} & Urban Traffic Forecasting \\ 
\bottomrule
\end{tabular}
\end{table*}

\subsection{Temporal Continual Learning}

Time series and temporal sequential data are commonly used data structures in dynamic systems. Such data are collected from sensors and continuously change with time, and this variation mainly manifests as data distribution shift.
There are several tasks for temporal sequential data, 
such as time series forecasting \cite{zhouInformerEfficientTransformer2021} (including long-term time series forecasting and short-term time series forecasting), 
time series imputation \cite{alcarazDiffusionbasedTimeSeries2022}, 
time series anomaly detection \cite{renTimeSeriesAnomalyDetection2019}, 
and time series classification \cite{yinContinualLearningAttentive2023}. Among them, forecasting tasks and imputation tasks are more likely to use generative or regression models, while anomaly detection and classification are more likely to use discriminative or classification models.

Deep neural network-based methods have been broadly developed for temporal sequence data mining for a long time. In the past few years, methods that based on Recurrent Neural Network(RNN) \cite{linSegRNNSegmentRecurrent2023, siami-naminiPerformanceLSTMBiLSTM2019, livierisCNNLSTMModel2020, yamakComparisonARIMALSTM2020} are the mainstream of temporal sequence data mining. Soon afterward, with the extreme success of Transformers \cite{vaswaniAttentionAllYou2017} in natural language processing and computer vision, a lot of work has explored applying the attention mechanism in Transformer to temporal sequence data mining \cite{zhouInformerEfficientTransformer2021, wenTransformersTimeSeries2023, liuITransformerInvertedTransformers2023}. Recently, MLP-based methods \cite{dasLongtermForecastingTiDE2023, chenTSMixerAllMLPArchitecture2023, ekambaramTSMixerLightweightMLPMixer2023} have also achieved good results in terms of accuracy and computational speed.

However, in real-world applications, data are collected over time, requiring repeated learning of new tasks. When learning incremental tasks, standard deep learning-based methods suffer from distribution shifts and catastrophic forgetting. Hence, modern models should be able to learn incremental knowledge continuously, and introducing the CL philosophy into temporal data mining is a promising direction. We thus propose a category named Temporal Continual Learning (TCL) to classify these works.

As a typical example of continuous time series forecasting, DoubleAdapt \cite{zhaoDoubleAdaptMetalearningApproach2023} design a two-fold adaptation, called data adaption and model adaption against distribution shift in stock price trend forecasting task. The data adaption transfers the feature distributions and posterior distribution to their corresponding agent distribution, which can close the gap between incremental data and test data. To optimize this two-fold adaptation, DoubleAdapt utilized ideal from MAML\cite {finnModelagnosticMetalearningFast2017}, which uses a two-step updating method to find an optimum in all tasks.

Continuous time series classification tasks have attracted more attention. The reason is that many well-studied CL strategies targeted classification tasks, and many class incremental learning methods from the computing version field can be migrated to continuous time series classification tasks.
For example, TTD \cite{yinContinualLearningAttentive2023} proposed a continual learning training method called Temporal Teacher Distillation. In the stage where the model learns the incremental task, TTD trains and tunes the model on each sequence of training datasets. To overcome catastrophic forgetting in LSTM, TDD proposed three hypotheses and combined them with distillation Loss and experience replay.
Gupta et al. \cite{guptaContinualLearningMultivariate2021} proposed a novel modularized neural network architecture to handle variable input dimensions, with two main modules: a core dynamics module comprising an RNN that models the underlying dynamics of the system, and a conditioning module using a GNN that adjusts the activations of the core dynamics module for each time-series based on the combination of sensors available, effectively exhibiting different behavior depending on the available sensors.
Dubey et al.\cite{dubeyContinualLearningTimetoEvent2022} proposed HyperHawkes, a sequence descriptor-conditioned hypernetwork-based neural Hawkes process that can generate sequence-specific parameters to address continual learning of event sequences. The neural Hawkes process comprises two building blocks - RNN and feedforward neural network (FNN) and overcomes forgetting by incorporating a regularization on the hypernetwork parameters such that it penalizes any change to the FNHP parameters produced from old sequences.




\subsection{Spatial-temporal Continual Learning}
Most data generated in cities come in spatial-temporal structures, which means that data not only have spatial features but also can change over time. The most critical challenge lies in their complex dependencies on various spatial and temporal indicators. Another practical issue is heterogeneity, meaning that the distribution drift can be not only in the time dimension but also in the space dimension. For these reasons, traditional methods that only focus on time or space, such as Auto-regressive Integrated Moving Average (ARIMA), Gradient Boosting Decision Tree (GBDT), ResNet, or AlexNet, can hardly tackle these problems completely. Fortunately, the recent advance in spatial-temporal data mining offers many spatial-temporal models, such as ConvLSTM, PredRNN, ST-ResNet, STGCN, and Graph WaveNet. Readers are referred to a few surveys \cite{wangDeepLearningSpatioTemporal2022, jinSpatioTemporalGraphNeural2023} to overview such frontiers.

However, even advanced models become limited when dealing with streaming data, incremental tasks, or new classes. It is because they mainly aim at handling one task at a certain time point, without considering any adaptation to future changes. For example, when a city expands, the model designed for the original urban structure may not adaptively capture the features of the new region; more importantly, the new region can produce a complex spatial-temporal dependence on the old region over time. 

To solve the problems above, recent works propose to apply continual learning techniques to spatial-temporal models. Kanazawa et al. \cite{kanazawaIncrementalLearningSpatialTemporal2019} used a Gaussian mixture model (GMM) to capture human motion patterns. They extracted spatial features from people's working states and temporal features from moving states. And finally, they implemented a new incremental learning algorithm for the GMM to adapt to new changes.
Xiao et al. \cite{xiaoShorttermTrafficVolume2019} proposed a regression to classification (R2C) framework that can train any ensemble algorithms on linear and non-linear SVRs. There are three steps in the proposed framework. The first step is to construct classification datasets from regression datasets; then, by integrating them with the proposed Learn++ algorithms, multiple classifiers can be integrated; finally, a regression model is constructed upon the classifiers, which is updated incrementally.

Trajectory data naturally contain spatial-temporal information. Bandaragoda et al. \cite{bandaragodaTrajectoryClusteringRoad2019} proposed an approach using hyper-dimensional computing to transform variable-length trajectories of commuter trips into fixed-length high-dimensional vectors, with the benefit of being incrementally learned over time.
For trajectory prediction, Ma et al.\cite{maContinualMultiAgentInteraction2021} proposed a graph-neural-network-based continual multi-agent trajectory prediction framework. It consists of three modules, a graph-neural-network-based predictor, an episodic memory buffer, and a conditional-variational-autoencoder-based generative memory module.

As for other applications, SARDINE \cite{dasSARDINESelfAdaptiveRecurrent2020} modeled the overall spatio-temporal prediction problem as the prediction of derived remote sensing imagery. To consider the influence of temporally evolving spatial features in the neighborhood of each pixel, SARDINE proposed a technique called layer growing for better modeling the spatial features' evolution in various spatial contexts. 
CSTWPP \cite{huVeryShorttermSpatial2020} was proposed to forecast wind power, and a CNN has been used for spatial-temporal feature extraction and wind power prediction. The whole training procedure of CSTWPP is divided into two steps: the first step is to train the CSTWPP model offline on the training set using SGD; and the second step is to train the CSTWPP online through incremental learning on the testing set, termed online training.
In WISDOM \cite{xuSpatioTemporalMultiTaskLearning2021}, the spatial-temporal data are assumed to be periodically augmented with a new data chunk. The goal was to adapt the existing models without rebuilding them from scratch when new observations were available. To ensure that the model parameters and latent factors do not vary significantly from their previous values, a smoothness criterion was added as a constraint to the objective function.
TF-Net \cite{yuIncrementalLearningBased2021} used incremental learning rather than batch learning to perform model training. Their traffic flow dataset was divided into multiple sub-train sets, and the model was iteratively trained with each set. When the model stopped to improve for a few epochs, it would be transferred to the next sub-train set for training. Similarly, IL-TFNet \cite{shaoTrafficFlowPrediction2021} proposed an incremental learning-based CNN-LTSM model, with a similar training phase.

TrafficStream \cite{chenTrafficStreamStreamingTraffic2021} is a typical spatial-temporal CL method, which used a simple GNN as a surrogate model for complex traffic flow forecasting methods. In TrafficStream, new nodes with their 2-hops neighbors were used to construct a sub-graph for mining the influence of network expansion. Further, an algorithm based on JS Divergence detected existing nodes whose traffic patterns changed significantly. To consolidate the previous knowledge, historical nodes of traffic networks were replayed with weighted smoothing constraints imposed on the current training model. 
To improve the TrafficStream, STKEC \cite{wangKnowledgeExpansionConsolidation2023} developed a new framework containing two major components. One was an influence-based knowledge expansion strategy, used for integrating newly evolved traffic patterns of expanding road networks; and the other was a memory-augmented knowledge consolidation mechanism, which was to consolidate the old knowledge of the previous road network.

\subsection{Multi-modal Continual Learning}

Various sensors deployed in cities can collect numerous types of urban information, recorded as vehicle trajectories, traffic flows, satellite data, user feedback, etc. These types of information are recorded as multi-modal heterogeneous data, including but not limited to images, videos, texts, maps, graphs, and point clouds. To this end, the trend of AI is transitioning from developing single-modal models to universal models that can process multiple modes of data. Thus multi-modal machine learning (MML) has emerged \cite{ramachandramDeepMultimodalLearning2017, baltrusaitisMultimodalMachineLearning2019, liangFoundationsTrendsMultimodal2023}.
In general, the objectives of MML include constructing a unified feature representation \cite{guoDeepMultimodalRepresentation2019}, achieving semantic alignment between different modes \cite{baltrusaitisMultimodalMachineLearning2019}, integrating and fusing the feature representation of different modes \cite{nagraniAttentionBottlenecksMultimodal2021}, generating new data of one mode based on another \cite{luoUniVLUnifiedVideo2020}, and transferring knowledge between modes \cite{vaezijozeMMTMMultimodalTransfer2020}. The downstream tasks of MML usually include visual question answering \cite{antolVQAVisualQuestion2015}, natural language for visual reasoning \cite{suhrCorpusReasoningNatural2019}, vision-language retrieval \cite{goenkaFashionVLPVisionLanguage2022}, and visual dialogue \cite{dasVisualDialog2017}.

Very recently, large-scale pre-training techniques \cite{devlinBERTPretrainingDeep2019, brownLanguageModelsAre2020} are prevailing. For example,  CLIP \cite{radfordLearningTransferableVisual2021} used natural language as a training signal to the image data, on the basis of a large dataset (400 million [image, text] pairs) and big backbone models (ResNet-50 \cite{heDeepResidualLearning2016} or ViT \cite{dosovitskiyImageWorth16x162020b}). Moreover, a simple comparative learning method \cite{zhangContrastiveLearningMedical2020} was added to make CLIP a multi-modal model with zero-shot capabilities, thus achieving good generalization on new unseen data. After CLIP, many large-scale pre-training multi-modal studies have been proposed \cite{wangLargescaleMultimodalPretrained2023, liMultimodalFoundationModels2023, yinSurveyMultimodalLarge2023}.

\begin{figure}[!t]
\centering
\includegraphics[width=3.4in]{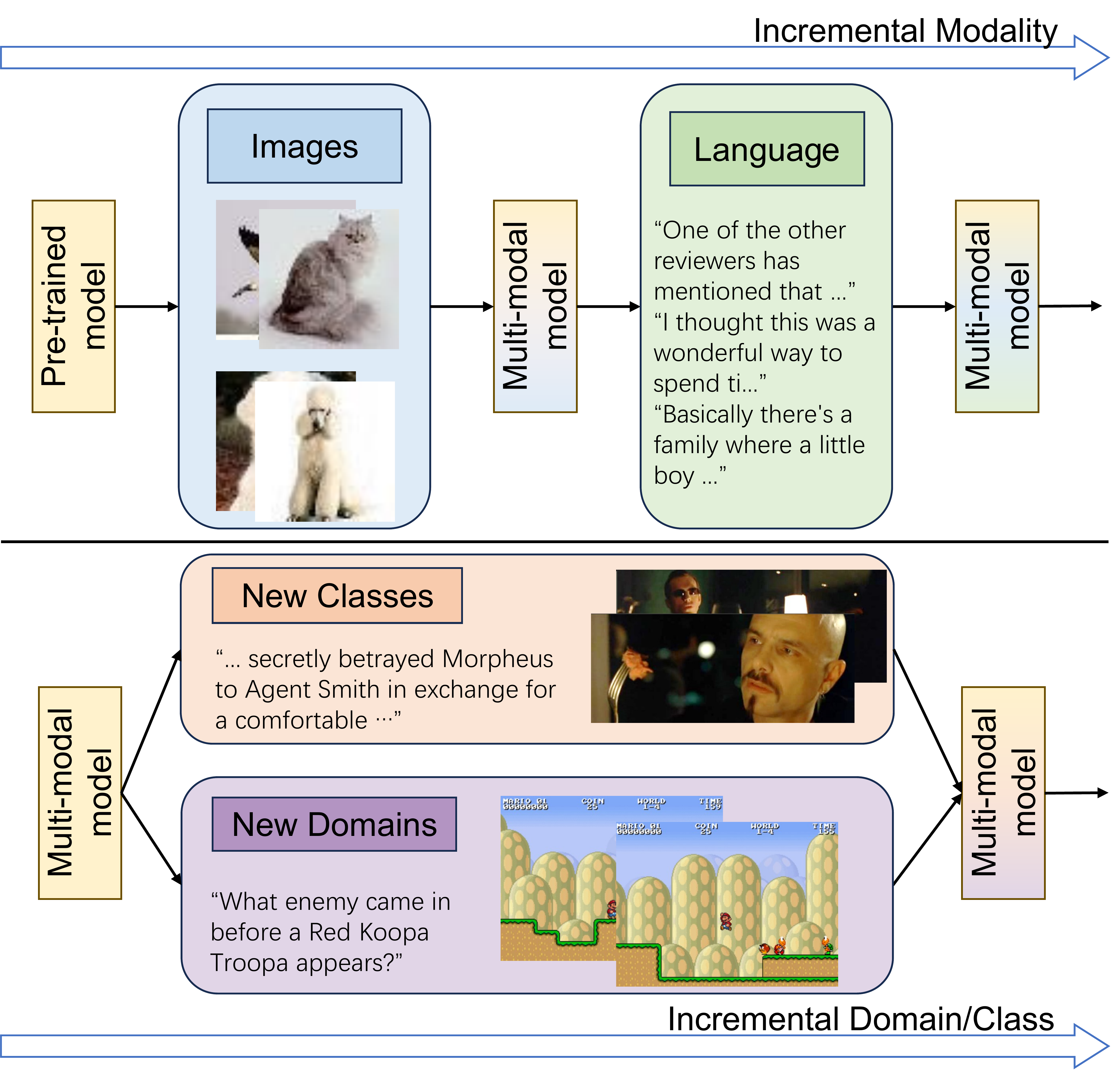}
\caption{A multi-modal continual learning (MCL) framework. In current literature, there are two types of tasks in MCL - one is modal incremental, and the other is the usual domain or class incremental. As shown in the upper part of the figure, the modal incremental problem requires the base model to continuously learn new modes of data. In the lower part, the other task aims to train the model with good generalization ability in multiple domains or in out-of-distribution situations.}
\label{fig_multi-modal}
\end{figure}

Despite the success, the catastrophic forgetting problem in many multi-task and multi-modal learning works has not been well solved. Therefore, multi-modal continual learning (MCL) has been proposed. An overview framework of MCL is shown in Fig. \ref{fig_multi-modal}.
As examples, CLiMB \cite{srinivasanCLiMBContinualLearning2022} introduced continual learning in multi-modality benchmark to facilitate the study of CL in vision-and-language tasks with deployment to multi-modal and uni-modal tasks. A learning problem was formulated where a model was first trained on sequentially arriving vision-and-language tasks, referred to as upstream continual learning, and then transferred downstream to low-shot multi-modal and uni-modal tasks.
Cai et al.\cite{caiMultimodalContinualGraph2022} presented a Multi-modal Structure-evolving Continual Graph Learning (MSCGL) model, which can adaptively explore model architectures without forgetting history information. MSCGL extracted multi-modal features using ViT and BERT in the data preprocessing stage, then designed different graph structures for different modalities, and used neural architecture search (NAS) to determine the new network architecture for searching the model with the best memory ability.



\subsection{Federated Continual Learning}
\begin{figure*}[!h]
\centering
\includegraphics[width=5in]{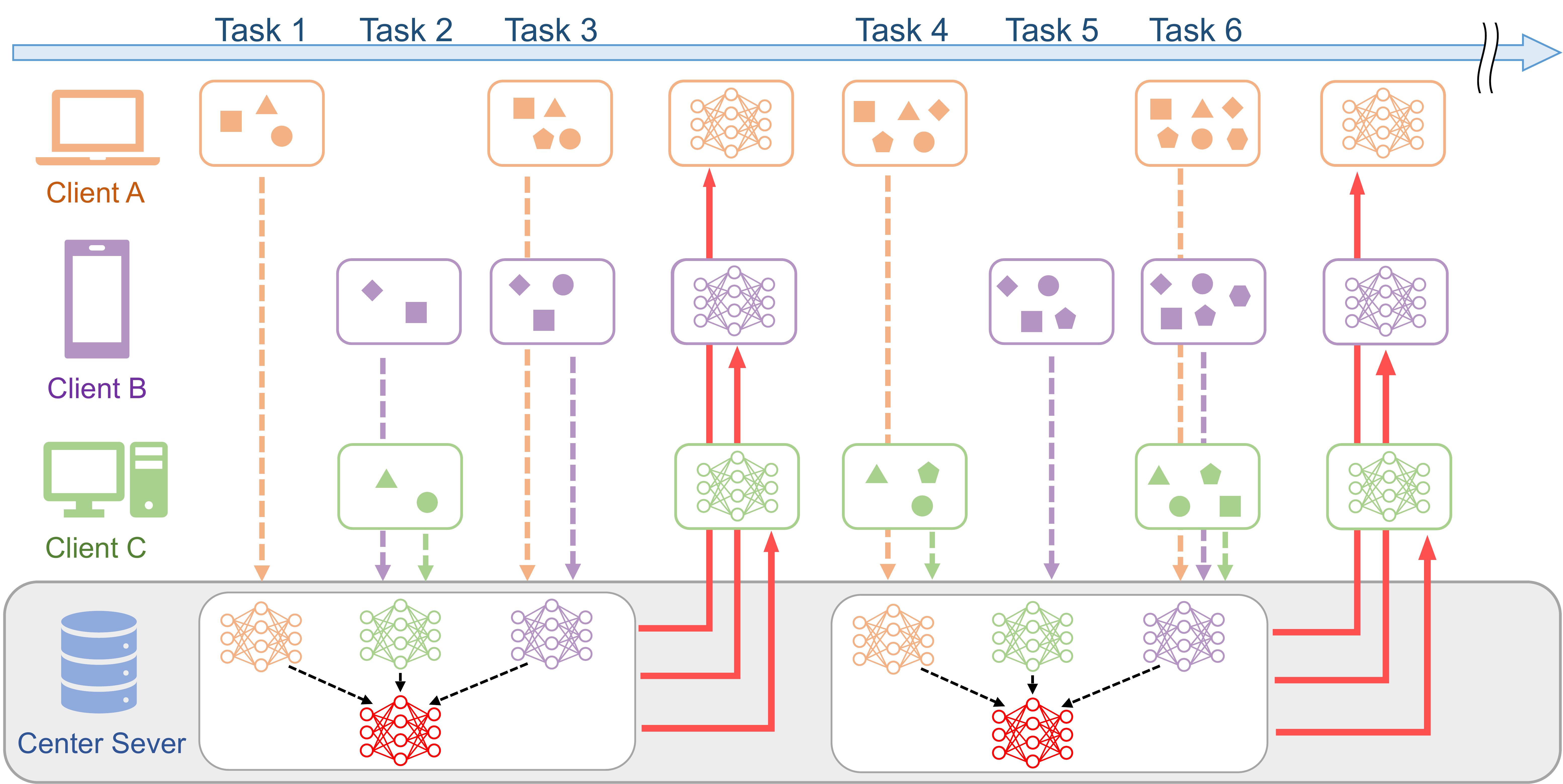}
\caption{A stander \textbf{federated continual learning} setting. Each client updates its local model on a sequence of tasks. For instance, client A updates a local model in tasks 1, 3, 4, and 6 with incremental samples or classes. Once updated, client A will transmit the parameters to the center server. Meanwhile, a global model is continually trained whenever new parameters arrive, and the updated parameters of the global model will be transmitted to each local model, marked in red line.}
\label{fig:FCL}
\end{figure*}

In reality, data are often generated from different sources. For example, traffic data come from traffic-management departments, and air quality data are from environment-protection departments. In addition, the Internet of Things (IoT), a new distributed technology, has facilitated the collection of data from numerous edge devices. Therefore, in modern situations, unified models need to process different data chunks from various sites in a decentralized way.
To this end, federated learning (FL) paradigm has been proposed to solve the challenges above \cite{konecnyFederatedLearningStrategies2017, huangFairnessAccuracyHorizontal2022, al-huthaifiFederatedLearningSmart2023}. By utilizing FL techniques, models can process different sources of data in a distributed way when facing data isolation and privacy problems.

Federated learning was first proposed by Macahan et al in 2017 \cite{mcmahanCommunicationEfficientLearningDeep2017}, where the classic FedAVG was presented to train distributed data simultaneously. More specifically, local data are first trained by local clients to construct initial models. Then, the models parameters' gradients are sent to the central server, and the global model's gradients are aggregated by weighted average. The FedAVG update is shown in Eq. \ref{eqn-fedavg}.  
\begin{equation}
    \label{eqn-fedavg}
    \omega_{t+1} = \sum_{k=1}^{K}\frac{m_k}{m}\omega_{t+1}^k
\end{equation}
where $\omega_{t+1}$ denotes the global parameters in the server, $\omega_{t+1}^k$ is the local parameters in the client $k$, and $\frac{m_k}{m}$ represents the proportion of client $k$ in the total sample of $M$ clients participating in training. Finally, the gradients of the updated global model are returned to the clients, and the above steps repeat until the global model converges.

More recently, federated learning has been combined with continual learning, the so-called federated continual learning (FCL) \cite{shohamOvercomingForgettingFederated2019, yoonFederatedContinualLearning2021}, to handle the challenges of dealing with incremental data or classes, non-stationary data distributions, and catastrophic forgetting problems that also happen in FL. A standard FCL setting is shown in Fig. \ref{fig:FCL}. 
Formally, Suppose a sequence of tasks with a center server $\mathcal{S}$ and a set of distributed clients $\mathcal{C}$. There are two goals of FCL - to minimize the average loss on all tasks and also to learn a global performer on all clients and the server. This can be written as Eq. \ref{eqn-FCL}.
\begin{equation}
    \label{eqn-FCL}
    \min_{\theta} \sum_{t = 1}^{T} \sum_{c \in \mathcal{C}} \frac{N_{c}^{t}}{N^{t}} \mathcal{L}(\mathcal{D}_{c}^{t} ; \theta )
\end{equation}
Where $\theta$ is model parameters, $t$ is the ID of each task, and $c$ is the ID of each client. $N_{c}^{t} / N^{t}$ represents the portion of training samples from client $c$ to all clients in task $t$. $\mathcal{D}_{c}^{t}$ is the training dataset of task $t$ and client $c$.

There are two main challenges in FCL, \textit{inter-client interference} and \textit{communication-efficiency}. The first challenge says that historical knowledge may decrease the performance of other clients, and the second one assumes that unlimited streams of tasks can cause intractable performance overhead. 
To mitigate the intra-task forgetting and inter-task forgetting problems in FCL, CFeD \cite{maContinualFederatedLearning2022} used a surrogate dataset that is publicly available and employed a distillation loss for clients to review old tasks and for the server to perform model aggregation. 
FedWeIT \cite{yoonFederatedContinualLearning2021} focused on the communication challenge. It separated \textit{general knowledge} and \textit{task-adaptive knowledge} by decomposing the local parameters into dense-base parameters and sparse task-adaptive parameters. 
FedSpace \cite{shenajAsynchronousFederatedContinual2023} proposed a new setting called Asynchronous Federated Continual Learning (AFCL), where each client has its own CL procedure till a communication step that updates all the clients to the server. To implement AFCL, FedSpace developed multiple procedures including Server Fractal Pre-training, Prototype Aggregation, Contrastive Representation Loss, and Server Aggregation.

The above FCL studies ignored some particularities of the distributed devices in urban environments. 
To alleviate catastrophic forgetting, FedSTIL \cite{zhangSpatialTemporalFederatedLearning2023} designed a lifelong learning framework for distributed edge clients. This edge framework included Extraction Layer, Adaptive Layer, and Prototype Rehearsal Storage. Specifically, the Extraction Layer was used for extracting raw data $ (X_i^{(t)}, Y_i^{(t)}) \in D_c^{(t)} $ for edge client $c$ on the $t$-th round to the extracted prototype set $ P_c^{(t)} $. To alleviate forgetting, the Prototype Rehearsal method was used to periodically sample and store some extracted prototypes of incremental tasks in local storage with the nearest-mean-of-exemplars strategy \cite{rebuffiICaRLIncrementalClassifier2017}. In the Adaptive Layer, the global knowledge from the server and local knowledge from clients are used for training by the following Eq. \ref{eqn-FedSTIL-theta} 
\begin{equation}
    \label{eqn-FedSTIL-theta}
    \theta_c = B_c \odot \alpha_c + A_c
\end{equation}
Where $A_c$ is the knowledge learned from local incremental tasks, $B_c$ is the base parameters with the spatial-temporal knowledge learned from global parameters, and $\alpha_c$ is the attention parameters to capture the task-specific knowledge.
 
Lastly, FpC \cite{lanzaUrbanTrafficForecasting2023} was proposed from the perspective of server-less FL based on the p2p paradigm \cite{huangContinualLearningPeertoPeer2022} and was the first to apply FCL to urban traffic flow prediction. In FpC, the model parameters were updated by going through all the edge clients with a random path. And by using FCL, the edge devices in smart cities enjoy many advantages like better privacy protection, lower communication overhead and latency, and smaller memory usage and energy consumption.

\section{Discussions}
\label{sec-discussion}
We have presented the basic and advanced methods, applications, and datasets of continual learning related to smart cities. Despite the remarkable advances, there remain quite many challenges to address. In this section, we discuss some of the important ones: continual large models, multi-model continual learning, continual learning in open world, privacy and security issues, and model explainability.

\subsection{Continual Large Models} 

Large Language Models (LLMs) such as GPT \cite{openaiGPT4TechnicalReport2023}, PaLM \cite{chowdheryPaLMScalingLanguage2022}, and LLaMA \cite{touvronLLaMAOpenEfficient2023} have demonstrated their strong learning intelligence and generalization.
The state-of-the-art LLMs have hundreds of billions of parameters, which makes them intractable to re-train or hard to update.

From the perspective of continual learning, different from traditional small pre-trained models, several critical issues hinder the evolution of LLMs. 
First, the primary limitation results from the high cost (computational, financial, time) of re-training or updating models. Therefore, more lightweight training methods and efficient CL strategies need to be devised. 
The second issue is attributed to completely different training approaches for LLMs. Often, new data for fine-tuning large models are dramatically small compared to the model size, so this inhibits large models from effectively acquiring new knowledge. 
The third challenge arises because LLMs are designed for strong generalization ability, likely on numerous domains. However, current CL methods primarily focus on maximizing knowledge within a single or few domains. Consequently, continual learning methods for LLMs must be particularly designed such that LLMs can learn extensive knowledge across various domains.
Furthermore, the absence of rigid analysis also poses a significant challenge to continual learning for LLMs. There is an urgent need to design insightful theories to guide the CL methods for LLMs. 
Lastly, ethical issues matter as well. LLMs must adhere to social morality and legal requirements, which can be achieved through methods such as alignment learning. This is a critical issue during the continual learning phase of LLMs.

In light of such problems, continual language learning (CLL) or continual language model (CLM) has started to emerge \cite{keContinualTrainingLanguage2022, jangContinualKnowledgeLearning2022, keContinualPretrainingLanguage2023, razdaibiedinaProgressivePromptsContinual2023}, as well as a survey on continual learning for LLMs \cite{wuContinualLearningLarge2024}.
As for smart city development, the main bottleneck lies in lacking well-developed large unified models. Many data in smart cities are generated in different forms like graphs, spatial-temporal, or distributed structures. However, the key technology such as large-scale pre-training techniques has not yet been adapted to such data structures. We believe that the trend would be to develop large unified models for processing different types of urban data. Afterward, how to continually update such large models would be an important future direction.

\subsection{Multi-modality Continual Learning}
In recent years, multi-modal learning is steadily developing and has flourished with large-scale pre-train models, such as CLIP \cite{radfordLearningTransferableVisual2021}, Stable Diffusion, BLIP-2 \cite{liBLIP2BootstrappingLanguageImage2023}, and multi-modality LLMs \cite{yinSurveyMultimodalLarge2023, liMIMICITMultiModalInContext2023, shenHuggingGPTSolvingAI2023, wuVisualChatGPTTalking2023, awadallaOpenFlamingoOpenSourceFramework2023}. 
In our view, smart city multi-modal CL faces the following challenges. The first one is the lack of high-quality and large-scale urban multi-modal datasets. In smart cities, such data are in fact rich, but because of many practical reasons such as privacy and security issues, obtaining such data has always been a major issue. Therefore, there is a great need for building more smart cities' multi-modal datasets and large-scale datasets. 
The second challenge lies in the continual learning part. Many pre-trained multi-modal models can be directly applied to smart cities, such as text, image, and video processing. However, when it comes to some data modes unique to urban computing, such as traffic and weather data, existing multi-modal models become insufficient, and performing only fine-tuning can hardly meet the actual needs. Some existing studies are limited to specific tasks and modes \cite{srinivasanCLiMBContinualLearning2022, caiMultimodalContinualGraph2022, gaoBMUMoCoBidirectionalMomentum2022}, so there is a large room for developing multi-modal models with continual learning abilities in smart city environments.

\subsection{Continual Learning in Open World}
In our view, continual learning in open worlds can be viewed from two perspectives: self-education and lifelong education. 
Self-education assumes that continual learning models must have the capability to acquire new knowledge \textit{autonomously}. That is, these models must independently detect new objects (OOD data or tasks) in the environment, understand, and convert them into memorized knowledge without human intervention or guidance. Ideally, such models are supposed to self-plan learning paths. 
As for lifelong education, we expect the model to possess the capability to explore new environments and integrate the new knowledge acquired into existing knowledge bases. Essentially, the model needs to continuously expand and adapt to environmental changes.
There are already studies on the concept of continual learning in open worlds, such as Continual Reinforcement Learning \cite{huangContinualModelBasedReinforcement2021, wolczykContinualWorldRobotic2021,khetarpalContinualReinforcementLearning2022, pengReinforcedIncrementalCrossLingual2023, abelDefinitionContinualReinforcement2023} and open-world continual learning \cite{liuAIAutonomySelfinitiated2023, truongFairnessContinualLearning2023, kimOpenWorldContinualLearning2023, liLearningPromptKnowledge2024}. However, these studies primarily concentrate on addressing the OOD problem, leaving much room of realizing the aforementioned objectives of self-education and lifelong education. Additionally, they didn't particularly focus on urban applications, where there are more intricate scenarios and demands.

\subsection{Privacy and Security}
As mentioned earlier, privacy and security issues have become one major obstacle hindering the development of unified models. 
On privacy, continual learning needs to collect user data periodically, but the rights may not always be authorized; furthermore, privacy leakage may occur everywhere during data collection, transmission, or storage processes. On security, distributed devices in cities are vulnerable to attack. If individual data are not securely protected, then data aggregation would become infeasible.

From one aspect, the security research community needs to define more comprehensive criteria for evaluating privacy and security attacks in real systems. Also, stronger protection techniques such as robust encryption algorithms should be developed, which are required for secure data collection and exchange. From another aspect, the continual learning community needs to research how intelligent models (central or distributed) can perform continual learning securely, especially when dealing with incomplete data pieces because of privacy or security constraints. We have seen works of federated continual learning that are devoted to this regard \cite{lanzaUrbanTrafficForecasting2023, zhangSpatialTemporalFederatedLearning2023, yoonFederatedContinualLearning2021, maContinualFederatedLearning2022, liPersonalizedFederatedContinual2023, decaroContinualAdaptationFederated2023, shenajAsynchronousFederatedContinual2023}. However, more complex and practical situations will motivate more secure continual learning scenarios and strategies.

Moreover, a significant security concern arises from continual learning itself. Due to the nature of CL, a learning model may have accumulated a vast amount of historical knowledge, which could potentially be illicitly extracted through various methods, leading to the leakage of critical information. Hence, there is a need to explore continual learning methods that can actively and selectively forget. Active and controlled forgetting is not only beneficial for enhancing the model's memory efficiency and performance \cite{qiFinetuningAlignedLanguage2023} but also important for safeguarding privacy and enhancing security protection.

\subsection{Explainability}
Explainable artificial intelligence (XAI) is one general trend in the machine learning community. Advanced models such as deep neural nets have significantly boosted the performance but also become much poorer in explainability, and many of their generated results are not interpretable.
For city management, the behaviors of any decision models before deployment must be well understood and interpreted by not only their developers but also the city administrators \cite{javedSurveyExplainableArtificial2023, ahmadDevelopingFutureHumancentered2022}.
However, the majority of smart city CL works have focused on overcoming catastrophic forgetting and improving predictive power by using much more complex models (such as expandable deep networks).  
A strong side effect would be that the explainability of such complex models becomes very unclear, and the study of explainability is no longer only a matter for basic models but also for their continual learning strategies. There is very little research on this regard. For example, ICICLE \cite{rymarczykICICLEInterpretableClass2023} proposed an interpretable class-incremental learning framework that can reduce the interpretability concept drift.
Hence, we believe that it is of great necessity to understand both a base model's explainability and also the rationale of its continual learning strategy.

\section{Conclusions}
\label{sec-conclusion}
We have presented a comprehensive survey of continual learning (CL) studies relevant to smart city research. We started with the basic task setting and scenarios of CL and then reviewed those widely used CL methods. Next, We categorized the primary applications and learning tasks of CL in smart cities and compiled a list of publicly available datasets commonly used in previous studies. We then offered analysis and examples showcasing the integration of CL with other learning frameworks, such as graph learning, temporal learning, spatial-temporal learning, multi-modality learning, and federated learning. Lastly, we outlined some important challenges faced by CL in smart cities and envisioned future directions. Urban computing and smart city research represent the most complex real-world applications, and we believe that our survey has covered the most fundamental literature on continual learning. We anticipate that this survey could help relevant researchers quickly familiarize themselves with the current state of CL and smart city research and then direct them to future research trends.


%

\appendices


\section*{Acknowledgments}

This work was supported in part by the National Natural Science Foundation of China (Grant No. 62302405, 62176221, 62276215), the Natural Science Foundation of Sichuan Province (Grant No. 24NSFC2348), China Postdoctoral Science Foundation (Grant No. 2023M732914), Sichuan Science and Technology Program (No. MZGC20230073) and the Fundamental Research Funds for the Central Universities (No. 2682023ZT007)

\ifCLASSOPTIONcaptionsoff
  \newpage
\fi


\bibliographystyle{IEEEtran}
\bibliography{bibtex/bib/IEEEabrv.bib,bibtex/bib/IEEEfull.bib}
%



\end{document}